\documentclass{article}

\usepackage{microtype}
\usepackage{graphicx}
\usepackage{booktabs} 
\usepackage{array} 
\usepackage{multirow} 
\usepackage{siunitx}
\usepackage{longtable}
\usepackage{stfloats}
\usepackage{caption}
\usepackage{subcaption}
\usepackage{amsthm}

\usepackage{hyperref}

\usepackage[accepted]{icml2025}

\usepackage{amsmath}
\usepackage{amssymb}
\usepackage{mathtools}
\usepackage{amsthm}

\usepackage[capitalize,noabbrev]{cleveref}

\theoremstyle{plain}

\theoremstyle{definition}

\theoremstyle{remark}

\usepackage[textsize=tiny]{todonotes}

\icmltitlerunning{SPACE: Your Genomic Profile Predictor is a Powerful DNA Foundation Model}

\begin{document}

\twocolumn[
\icmltitle{SPACE: Your Genomic Profile Predictor is a Powerful DNA Foundation Model}
% Supervised Genomic Profile Prediction as a Stronger DNA Representation Learner}
%Is Unsupervised Pre-training Necessary for DNA Foundation Models? Supervised Genomic Profile Learning as a Stronger DNA Representation Approach
%Rethinking DNA Foundation Models: Supervised Genomic Profile Prediction as a Superior Alternative to Unsupervised Pre-training



\icmlsetsymbol{equal}{*}

\begin{icmlauthorlist}
\icmlauthor{Zhao Yang}{equal,yyy,zzz}
\icmlauthor{Jiwei Zhu}{equal,yyy,zzz}
\icmlauthor{Bing Su}{yyy,zzz}
\end{icmlauthorlist}

\icmlaffiliation{yyy}{Gaoling School of Artificial Intelligence, Renmin University of China, Beijing, China}
\icmlaffiliation{zzz}{Beijing Key Laboratory of Research on Large Models and Intelligent Governance}

\icmlcorrespondingauthor{Bing Su}{bingsu@ruc.edu.cn}

% the "keywords" metadata in the PDF but will not be shown in the document#
\icmlkeywords{Machine Learning, ICML}

\vskip 0.3in
]



\printAffiliationsAndNotice{\icmlEqualContribution} % otherwise use the standard text.

\begin{abstract}
Inspired by the success of unsupervised pre-training paradigms, researchers have applied these approaches to DNA pre-training. However, we argue that these approaches alone yield suboptimal results because pure DNA sequences lack sufficient information, since their functions are regulated by genomic profiles like chromatin accessibility. Here, we demonstrate that supervised training for genomic profile prediction serves as a more effective alternative to pure sequence pre-training. Furthermore, considering the multi-species and multi-profile nature of genomic profile prediction, we introduce our \textbf{S}pecies-\textbf{P}rofile \textbf{A}daptive \textbf{C}ollaborative \textbf{E}xperts (SPACE) that leverages Mixture of Experts (MoE) to better capture the relationships between DNA sequences across different species and genomic profiles, thereby learning more effective DNA representations. Through extensive experiments across various tasks, our model achieves state-of-the-art performance, establishing that DNA models trained with supervised genomic profiles serve as powerful DNA representation learners. The code is available at \url{https://github.com/ZhuJiwei111/SPACE}.
% compared to unsupervised pre-training approaches.

%Inspired by the success of unsupervised learning paradigms in Natural Language Processing (NLP), researchers have attempted to apply similar pre-training approaches like BERT and GPT to DNA sequences, often referred to as the "language of life." However, we argue that direct pre-training on DNA sequences alone fails to achieve the anticipated success, primarily due to the insufficient information content of raw DNA sequences, as their functional roles in organisms are regulated by multiple factors such as cell types and epigenetic modifications. In this paper, we demonstrate that supervised training for genomic profile prediction serves as a more effective alternative to pure sequence pre-training. Furthermore, considering the multi-task and multi-species nature of DNA genomic profile prediction, we introduce our dual-layer Mixture of Experts (MoE) design to better capture the relationships between DNA sequences across different profiles and species. Through extensive experiments across various tasks, our model achieves state-of-the-art (SOTA) performance, establishing that DNA models trained with supervised genomic profiles serve as more powerful DNA representation learners compared to traditional unsupervised pre-training approaches.
\end{abstract}

\section{Introduction}

DNA sequences, encoded by four nucleotide bases (\texttt{A}, \texttt{C}, \texttt{G}, and \texttt{T}), serve as biology's fundamental language that carries genetic instructions. Understanding the syntax and regulatory grammar of this molecular language is promising for diverse applications, including disease diagnosis~\citep{kernohan2024expanding, sermon2004preimplantation}, drug discovery~\citep{peterson2023small, trajanoska2023target}, and protein engineering~\citep{gosai2024machine, yang2025regulatory}. 

Due to the complexity of DNA sequences, gaining a clear understanding of DNA is not easy. Inspired by the success of unsupervised pre-training paradigms in NLP, such as masked language modeling~\citep{devlin2019bert} (MLM) and next-token prediction~\citep{brown2020language} (NTP), several DNA foundation models (DFMs) have recently emerged following similar pre-training approaches. DFMs~\citep{DNABert, DNABert2, NT, nguyen2024sequence, HyenaDNA} aim to learn transferable representations by leveraging large-scale pre-training on massive DNA sequences, which can be adapted for exploring the functions and mechanisms of DNA. Through fine-tuning, these models have demonstrated promising performance in tasks such as regulatory element identification, splice site recognition, and epigenetic modification prediction.

Although DFMs have made some progress, we argue that solely applying unsupervised pre-training techniques to DNA sequences alone cannot learn high-level semantic representations with strong generalization capabilities~\citep{tang2023current, tang2024evaluating}. Natural language sequences themselves can express their meaning relatively completely. In contrast, the functional roles of DNA sequences in organisms are far more complex and are regulated by numerous genomic profiles in a cell-type-specific manner~\citep{fu2025foundation}, including various epigenetic modifications~\citep{portela2010epigenetic}, chromatin accessibility~\citep{tan2023cell}, and transcription factor binding~\citep{peterson2023small}. Therefore, without incorporating these additional biological contexts, DFMs pre-trained only on raw DNA sequences may struggle to generalize to diverse cell environments or provide meaningful insights into the underlying biology.

Given that DNA's functional roles are regulated by various biological factors beyond sequence alone, we revisit supervised genomic profile prediction models (GPPMs) as an alternative to unsupervised DFMs for learning DNA sequence representations. These models~\citep{zhou2015predicting, kelley2018sequential, zhou2018deep, chen2022sequence, enformer} are trained to predict experimentally measurable genomic profiles that directly encode regulatory and functional information in a cell-type-specific manner. By learning to map sequences to these biologically meaningful profiles that reflect the complex regulatory mechanisms described above, supervised models may capture more functionally relevant representations compared to unsupervised DFMs that rely solely on sequence patterns. Such biological context-aware representations could be better aligned with downstream genomic applications, such as identifying regulatory elements or understanding gene expression patterns. While \citet{NT} has demonstrated through preliminary experiments that Enformer~\citep{enformer} can learn effective DNA representations, there has been no systematic study on improving the representation learning capabilities of GPPMs.

However, current GPPMs employ oversimplified architectures, using a shared encoder for DNA sequences from different species and independent prediction heads for different genomic profiles. This design has two major limitations. First, the species-shared encoder fails to capture species-specific characteristics, as regulatory mechanisms and their influences often vary across species~\citep{SpeciesLM}. These distinct features are crucial for understanding subtle genomic variations and context-dependent expression patterns. Second, genomic profile prediction inherently involves multiple interrelated tasks~\citep{fu2025foundation}, as different profiles influence each other and are often regulated by common mechanisms. The independent prediction heads, however, prevent the model from capturing these cross-profile dependencies and their variations across species.

To effectively model both cross-species and cross-profile relationships, we introduce our \textbf{S}pecies-\textbf{P}rofile \textbf{A}daptive \textbf{C}ollaborative \textbf{E}xperts (SPACE), which consists of two key components: (1) a species-aware encoder module and (2) a profile-grouped enhancement decoder module, both built upon Mixture of Experts (MoE). The species-aware encoder employs sparse routing to dynamically balance species-specific and shared biological features, while the profile-grouped enhancement decoder uses dual-gated expert weighted aggregation to capture the intricate dependencies between different genomic profiles. This design enables our model to effectively learn both species-specific patterns and shared regulatory mechanisms across profiles. 

The major contributions of this paper include:
\begin{itemize} 
\item We revisit the supervised pre-training paradigm for DNA sequence foundation models through genomic profile prediction as the pre-training objective, demonstrating how function-related biological contextual information can be effectively encoded into the learned representations.

\item We propose SPACE, a novel architecture that leverages MoE to better capture the relationships between DNA sequences across different species and genomic profiles, thereby learning more effective DNA representations compared to current GPPMs, which simply employ a shared encoder and independent prediction heads.

\item Through extensive experiments across a wide range of tasks, our SPACE achieves state-of-the-art (SOTA) performance, demonstrating that supervised pre-training for genomic profile prediction serves as a more effective and powerful alternative to pure sequence pre-training.
\end{itemize}
\section{Related Work}

\textbf{Supervised genomic profile models} are trained to predict functional genomic profiles from DNA sequences~\citep{kathail2024leveraging}. DeepSEA~\citep{zhou2015predicting} pioneered this paradigm by leveraging convolutional neural networks (CNNs) to extract DNA sequence features for multi-task prediction. Subsequent works~\citep{kelley2018sequential, zhou2018deep, chen2022sequence} have continued to advance this direction through either more advanced architectures or larger-scale training data. Enformer~\citep{enformer}, widely recognized as the SOTA method, achieved superior prediction performance through a hybrid Transformer-CNN architecture. While these methods primarily focus on \textit{ab initio} prediction of genomic profiles from DNA sequences and directly utilize these profiles for downstream tasks such as variant effect prediction, few studies~\citep{NT} have explored whether their intermediate representations capture meaningful biological patterns. Moreover, these models, which typically adopt a shared encoder coupled with independent profile prediction heads, have not thoroughly explored more effective architectural designs that could potentially enhance both prediction performance and representation learning.

\textbf{Unsupervised DNA foundation models} draw from the success of unsupervised pre-training in NLP. DNABERT~\citep{DNABert} pioneered this approach, maintaining nearly identical training methods to BERT~\citep{devlin2019bert} while adapting the tokenization scheme to 6-mers~\citep{celikkanatrevisiting} for DNA sequences. Subsequent works have continued along this direction, employing either MLM~\citep{DNABert2, NT, sanabria2024dna} or NTP~\citep{nguyen2024sequence, HyenaDNA} as unsupervised training objectives. Although these methods have made effective optimizations in terms of training data, model architectures, and tokenization strategies, they still adhere to the assumption that unsupervised pre-training on pure DNA sequences alone is sufficient for learning effective representations. Moreover, there has been little systematic comparison between these models and genomic profile prediction models in terms of their representation learning capabilities.

\textbf{The MoE framework} is a conditional computation technique that selectively activates different expert networks for different inputs through sparse routing~\citep{MoE0, SparseMoE}. In Transformer-based large language models (LLMs), MoE is typically applied to feed-forward networks (FFNs) to achieve better parameter efficiency while maintaining model capacity~\citep{fedus2022switch, jiang2023mistral, deepseek}. This adaptive routing mechanism is particularly well-suited for our genomic modeling task, as it enables the model to dynamically balance between learning species-specific patterns and shared biological features, while also capturing the complex dependencies between different genomic profiles. Following common practice in Transformer architectures, we also implement MoE by replacing the FFNs in our model.

\begin{figure*}[ht]
\vskip 0.2in
\begin{center}
\centerline{\includegraphics[width=\textwidth]{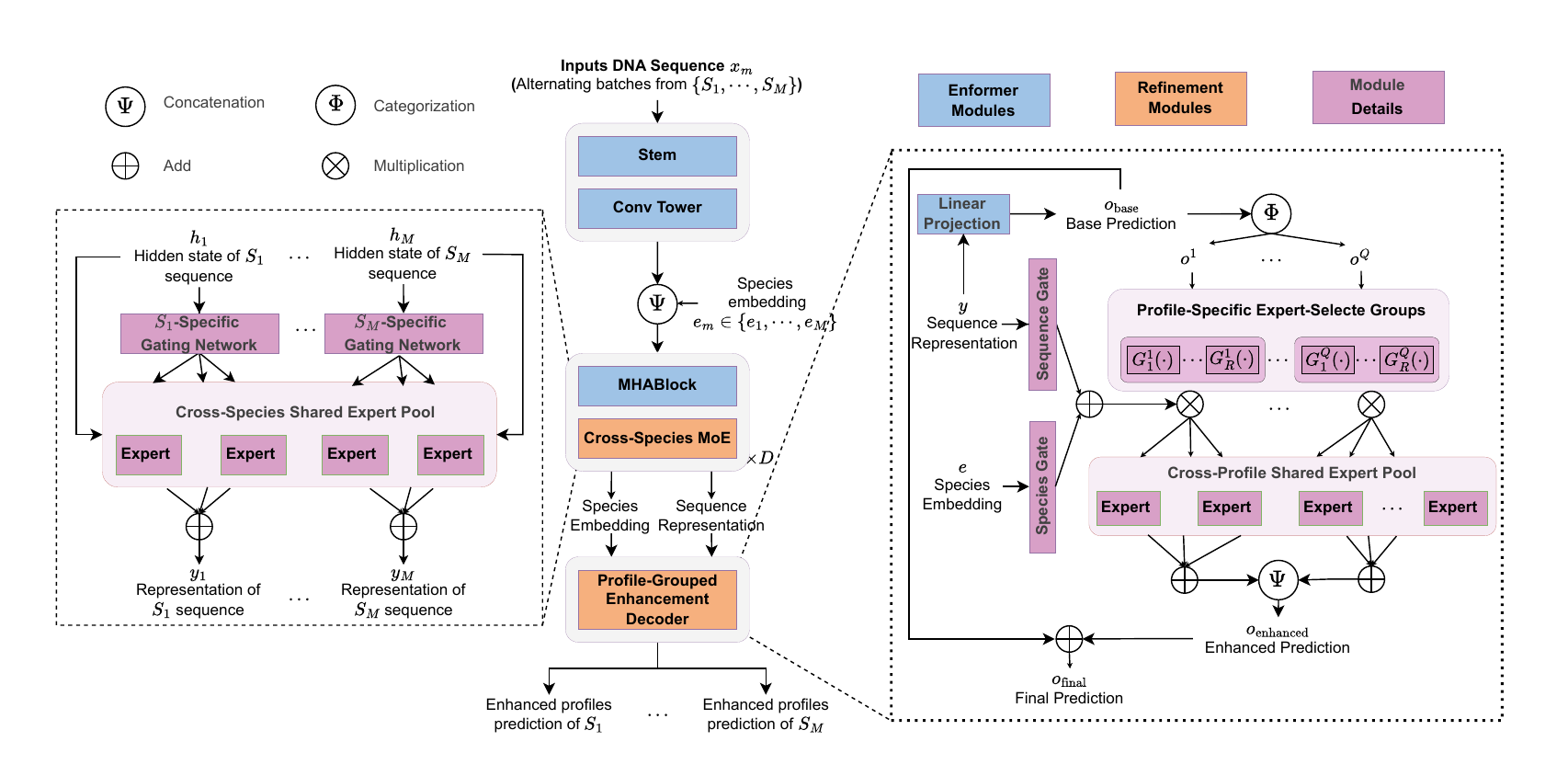}}
\caption{\textbf{Overview of our SPACE architecture.} It processes the input DNA sequence with three stages: (1) spatial compression and local context aggregation via a CNN-based aggregation module; (2) latent representation learning via a species-aware sparse MoE-based encoding module; (3) multi-profile prediction decoder via the dual-gated expert weighted prediction enhancement module. The detailed structures of the encoding module and the dual-layer gated prediction enhancement module are shown in the left and right, respectively.}
\label{model}
\end{center}
\vskip -0.2in
\end{figure*}

\section{Method}
\subsection{Problem Formulation}
Consider DNA sequences from $M$ species $\{S_1, \dots, S_M\}$. For each sequence $x_m$ from species $S_m$, we predict $C_m$ genomic profile values. We train with interleaved batches across all $M$ species to facilitate cross-species knowledge transfer~\citep{basenji2, enformer}. Through this supervised pre-training, the learned representations are expected to capture rich biological and regulatory information.

\subsection{Overview}
To better capture cross-species and cross-profile representations, we present SPACE. As illustrated in \cref{model}, our architecture consists of three key stages: (1) CNN-based Local Context Aggregation following Enformer~\citep{enformer}; (2) Species-aware Transformer Encoder and (3) Profile-Grouped Enhancement Decoder for genomic profile prediction. 

\subsection{Local Context Aggregation}
Given an input DNA sequence $x_m$, we first follow Enformer~\citep{enformer} to compress and aggregate the raw nucleotides through 1D-CNNs, generating hidden states $h_m \in \mathbb{R}^{L\times d_h}$ at 128bp resolution, where $L$ denotes the compressed sequence length and $d_h$ is the hidden dimension.

\subsection{Species-aware Encoder}
\label{sec:species_encoder}

Previous approaches to cross-species modeling~\citep{basenji2, enformer} typically employ a shared encoder for all species, lacking fine-grained modeling of species relationships. To address this limitation, we propose a novel cross-species modeling framework consisting of Species-specific Embedding and Cross-species MoE layers.

\textbf{Species-specific embedding.} We augment the aggregated hidden states $h_m$ with a trainable species-specific embedding $e_m \in \mathbb{R}^{1 \times d_h}$ by concatenation. The combined representation then passes through $D$ transformer layers with our Sparse Cross-species MoE for further transformation. This design is analogous to the source tokens used in recent language models~\citep{jiang2023mistral}, where document-level embeddings are prepended to provide explicit context about the content source. In our case, the species-specific embedding serves as an explicit signal to guide the model in distinguishing and handling species-specific characteristics. Among DNA models, ~\citet{SpeciesLM} employed a similar approach.

\textbf{Cross-species MoE.}
Furthermore, we introduce a sparse MoE encoding module that enables adaptive species-aware representation learning through dynamic parameter routing. For the $M$ species $\{S_1,...,S_M\}$, each MoE layer consists of two core components:
(1) a set of $N$ shared expert networks $\{E_1,...,E_N\}$, and
(2) $M$ species-specific gating networks $\{G_1,...,G_M\}$, where each $G_m$ is associated with species $S_m$ to dynamically weight expert contributions based on species-specific patterns.

For an aggregated hidden state $h_m$ from species $S_m$, the output representation $y_m$ is computed as:  
\begin{equation}
\begin{aligned}
\hat{h}_m &= \text{MHAttention}([e_m, h_m]),\\
y_m &= \sum_{k=1}^N \underbrace{G_m(\hat{h}_m)_k}_{\text{the k-th value of }G_m(\hat{h}_m)} \cdot E_k\left( \hat{h}_m \right),
\end{aligned}
\end{equation}  
where $e_m \in \mathbb{R}^{d_h}$ denotes the species embedding vector, and $[\cdot]$ represents concatenation, $\hat{h}_m$ is the hidden state after attention. The gating weights are computed through:  
% \begin{equation}
% G_m(c) = \text{softmax}\left(\text{TopK}\left(g(c) + \mathcal{R}_{\text{noise}}\right)\right),
% \end{equation}  
\begin{equation}
\label{eq:species_gating}
G_m(\hat{h}_m) = \text{Softmax}\left(\text{TopK}\left(g(\hat{h}_m) + \mathcal{R}_{\text{noise}}\right)\right),
\end{equation}
where $g(\cdot)$ is the gating function and $\mathcal{R}_{\text{noise}}$ is a noise injection term for training stability enhancement~\citep{fedus2022switch}.
% Moreover, to explicitly guide expert networks in learning both conserved and species-specific patterns, we introduce an expert-species mutual information loss inspired by Mod-Squad~\citep{Mod-Squad}:
Moreover, to explicitly guide expert networks in learning both conserved and species-specific patterns, we introduce an expert-species mutual information loss inspired by Mod-Squad~\citep{Mod-Squad}. We maximize the mutual information between species identity $S$ and expert selection $E$ to encourage species-specific expert specialization:
\begin{equation}
\begin{aligned}
\label{eq:MI}
\mathcal{L}_{\text{MI}} &= - MI(S;E) = -H(S) - H(E) + H(S,E) \\
&= \sum_{i=m}^M P(S_m)\log P(S_m) + \sum_{n=1}^N P(E_n)\log P(E_n) \\
&\quad - \sum_{m=1}^M\sum_{n=1}^N P(S_m,E_n)\log P(S_m,E_n),
\end{aligned}
\end{equation}
where $S_m$ denotes the species probability and $E_n$ represents the selection weight of each expert. The detailed derivations are provided in \cref{appendix:MI}.

After the encoding stage, we obtain the sequence representation $y \in \mathbb{R}^{L \times d_h}$ that captures both species-specific and shared biological features. Note that the sequence length remains $L$ as we exclude the species embedding token from the final representation, retaining only the original sequence positions for downstream genomic profile prediction.

% \subsection{Profile-Grouped Enhancement Decoder}
% Current GPPMs treat profile prediction as independent multi-tasks, overlooking inherent relationships between genomic profiles. This approach neglects two key biological principles: (1) evolutionary conservation implies shared regulatory mechanisms across homologous profiles in different species~\citep{schmidt2010five} and (2) different genomic profiles often share regulatory mechanisms and exhibit mutual influences~\citep{fu2025foundation}. To leverage these biological insights, we propose a dual-layer gated prediction enhancement module that enables systematic knowledge sharing across profiles. For clarity, we present the formulation for a single species $S_m$ and omit the subscript $m$ in subsequent notation.
\subsection{Profile-grouped Enhancement Decoder}

Current GPPMs treat profile prediction as independent multi-tasks, overlooking inherent relationships between genomic profiles. This limitation motivates our decoder design based on the following biological principles:

\textbf{Biological principles for profile relationships.} Genomic profile prediction should account for two key biological relationships: \textbf{(P1)} \textit{Evolutionary conservation}: Shared regulatory mechanisms exist across homologous profiles in different species~\citep{schmidt2010five}. \textbf{(P2)} \textit{Functional interdependencies}: Different genomic profiles often share regulatory mechanisms and exhibit mutual influences, e.g., regions with high gene expression typically exhibit increased chromatin accessibility~\citep{fu2025foundation}.

To leverage these biological insights, we propose a dual-layer gated prediction enhancement module that enables systematic knowledge sharing across profiles. For clarity, we present the formulation for a single species $S_m$ and omit the subscript $m$ in subsequent notation.

\textbf{Genomic profiles} can be categorized based on their experimental assays: for instance, DNase and ATAC-seq measure chromatin accessibility, while CAGE quantifies gene expression levels. Profiles from the same experimental type typically share similar functional mechanisms, enabling knowledge transfer within each category. Given $Q$ distinct profile types $\{T_1,..., T_Q\}$ with specific biological interpretations, for the DNA sequence representation $y \in \mathbb{R}^{L \times d_h}$ and the species embedding $e \in \mathbb{R}^{d_h}$, the enhancement module operates through the following sequential steps.

\textbf{Profile categorization for initial predictions.}
We first perform a linear projection on $y$ to obtain the initial base prediction $o_{\text{base}}$, which represents the final profile predictions from previous GPPMs~\citep{basenji2, enformer} that do not incorporate biological insights. Based on biological priors, $o_{\text{base}}$ is categorized into $Q$ independent parts $\{o^1, \dots, o^Q\}$, as follows. 
\begin{equation}
\begin{aligned}
&o_{\text{base}} = (\text{Linear}(y))^T \quad \in \mathbb{R}^{d_{\text{out}}\times L},\\ 
&\{o^1, \dots, o^Q\} = \Phi(o_{\text{base}}),
\end{aligned}
\end{equation}
where $d_{\text{out}}$ denotes the dimension specifying the total number of genomic profiles (i.e., $d_{\text{out}}$ equals $C_m$ for species $S_m$). The category operator $\Phi(\cdot)$ is constructed based on domain-specific biological knowledge, which decomposes the base prediction into $Q$ profile types $\{o^q\}_{q=1}^Q$ where $o^q \in \mathbb{R}^{d_q\times L}$ corresponds to biological profile type $T_q$, with $d_q$ indicating the number of profiles categorized to $T_q$.

\textbf{Dual-gated expert weighted aggregation.}
Each dimension of $o^q$ represents the base predicted sequence for a specific profile track. To capture shared regulatory patterns, we employ $K$ cross-profile-type shared experts $\{E_k\}_{k=1}^K$, where each expert $E_k: \mathbb{R}^{d_q \times L} \to \mathbb{R}^{d_q \times L}$ enhances the categorized base prediction $o^q$.

For adaptive expert selection, we introduce a dual-gated mechanism with two complementary components. \textbf{The first layer (group-level gating)} addresses principle \textbf{(P1)} by dynamically assigning weights to  $R$ expert groups based on species and sequence context, capturing evolutionary conservation patterns. The group weights $\hat{G}^q$ are computed as:
\begin{equation}
\hat{G}^q = \text{Softmax}\left( G^q_{\text{species}}(e) + G^q_{\text{sequence}}(\text{Pool}(y)) \right),
\end{equation}
where $G^q_{\text{species}}(\cdot)$ and $G^q_{\text{sequence}}(\cdot)$ map $\mathbb{R}^{d_h} \to \mathbb{R}^R$, weighting the $R$ expert-selected groups from species and sequence perspectives, respectively. Here, $\text{Pool}(\cdot)$ denotes global average pooling applied along the sequence length $L$ and $q$ indicates the profile type. \textbf{The second layer (expert-level gating)} addresses principle \textbf{(P2)} by selecting specific experts based on profile prediction patterns, capturing functional interdependencies between profiles. For each group $r$ and profile type $q$, we define the expert selection function $G_r^q(o^q) \in \mathbb{R}^K$, where $G_r^q(o^q)_k$ represents the weight of the $k$-th expert within the $r$-th group, determined by the base prediction patterns. The computation of $G_r^q$ follows a similar gating mechanism as in Equation~\ref{eq:species_gating}.

Combining both gating layers, the enhanced prediction for profile type $T_q$ is formulated as:
\begin{equation}
\begin{aligned}
o^q_{\text{enhanced}} = \sum_{r=1}^R \underbrace{\hat{G}_r^q}_{\text{Group weight}} \cdot \left( \sum_{k=1}^K \underbrace{G_{r}^q(o^q)_k}_{\text{Expert weight}} \cdot E_k(o^q) \right),
\end{aligned}
\end{equation}
where the dual-gated mechanism first weights relevant expert groups through $\hat{G}_r^q$ (based on species and sequence context), then within each selected group, chooses appropriate experts through $G_r^q(o^q)_k$ (based on prediction patterns).

The final predictions are computed through connections between enhanced and base predictions:
\begin{equation}
\begin{aligned}
o_{\text{enhanced}} &=\Psi\left( \{o^1_{\text{enhanced}}, ..., o^Q_{\text{enhanced}}\}\right), \\
o_{\text{final}} &= o_{\text{base}} + o_{\text{enhanced}}^T,
\end{aligned}
\end{equation}
where $\Psi(\cdot)$ is the inverse operator of $\Phi(\cdot)$, concatenating the enhanced predictions across different profile types, and the residual connection ensures stable training.

\subsection{Training Objective}

Following Enformer~\citep{enformer}, we adopt the Poisson negative log-likelihood as the primary loss function. To further refine species-aware expert selection in \cref{sec:species_encoder} by maximizing mutual information between species proportion and expert activations, we introduce an auxiliary mutual information loss. The composite loss is defined as:
\begin{equation}
\mathcal{L}_{\text{total}} = \mathcal{L}_{\text{Poisson}} - \alpha \sum_{d=1}^D MI(S;E_d),
\end{equation}

where $\alpha=0.01$ controls the mutual information regularization strength, $D$ denotes the number of transformer layers, $S$ represents the species identifier, and $E_d$ indicates the shared expert pool at layer $d$, the Poisson loss $\mathcal{L}_{\text{Poisson}}$ is mathematically formulated in \cref{sec:poisson}.

\section{Experiments}

\subsection{Experiment Setup}

\textbf{Pre-training dataset.} The pre-training dataset aligns with that used in Enformer~\citep{basenji2, enformer}, containing DNA sequences and corresponding genomic profiles for human and mouse genomes. Both species shared four conserved profile types: chromatin accessibility (DNase/ATAC-seq), transcription factor binding (TF ChIP-seq), histone modifications (Histone ChIP-seq), and transcriptional activity (CAGE). The number of profiles varies among different profile types in different species, with detailed dataset specifications provided in \cref{appendix:Pre-training Dataset}.

\textbf{Implementation details.}~Our model was pre-trained using supervised genomic profile prediction, maintaining the same prediction targets and genomic intervals as implemented in Enformer~\citep{enformer}. For cross-species joint modeling, we implemented an alternating training strategy using eight NVIDIA A40 GPUs. Training proceeded for 50,000 steps (approximately 8 days) with a global batch size of 64, achieved through 8 gradient accumulation steps (1 sample per GPU). Optimization employed AdamW~\citep{adamw} with an initial learning rate of $5 \times 10^{-4}$, linearly ramped from 0 during the first 5,000 steps followed by cosine decay. Gradient norms were clipped at 0.2 to maintain stability.

\begin{table*}[ht]
\caption{\textbf{MCC performance of Nucleotide Transformer downstream tasks.} This benchmark includes three categories of downstream tasks, comprising a total of 18 datasets derived from human samples. The term `NT downstream tasks' will be used to refer to these tasks.}
\vskip 0.15in
\centering
\label{tab:NT_small}
\resizebox{\textwidth}{!}
{ % 缩放表格至页面宽度
\begin{tabular}{l*{6}{S[table-format=2.2]}}
\toprule

\cmidrule(lr){1-7}
\multirow{2}{*}{Model} & 
\multicolumn{6}{c}{\textbf{Chromatin profiles}}\\
\cmidrule(lr){2-7}
& {H2AFZ} & {H3K27ac} & {H3K27me3} & {H3K36me3} & {H3K4me1} & {H3K4me2} \\
\cmidrule(lr){1-7}
DNABERT-2 &{0.490 $\pm$ 0.013} &{0.491 $\pm$ 0.010} &{0.599 $\pm$ 0.010} &\textbf{0.637 $\pm$ 0.007} &{0.490 $\pm$ 0.008} &{0.558 $\pm$ 0.013} \\
HyenaDNA-32KB &{0.467 $\pm$ 0.012} &{0.421 $\pm$ 0.010} &{0.550 $\pm$ 0.009} &{0.553 $\pm$ 0.011} &{0.423 $\pm$ 0.016} &{0.515 $\pm$ 0.018} \\
NT-HumanRef (500M) &{0.465 $\pm$ 0.011} &{0.457 $\pm$ 0.010} &{0.589 $\pm$ 0.009} &{0.594 $\pm$ 0.004} &{0.468 $\pm$ 0.007} &{0.527 $\pm$ 0.011} \\
NT-1000G (500M) &{0.464 $\pm$ 0.012} &{0.458 $\pm$ 0.012} &{0.591 $\pm$ 0.007} &{0.581 $\pm$ 0.009} &{0.466 $\pm$ 0.006} &{0.528 $\pm$ 0.011} \\
NT-1000G (2.5B) &{0.478 $\pm$ 0.012} &{0.486 $\pm$ 0.023} &\textbf{0.603 $\pm$ 0.009} &{0.632 $\pm$ 0.008} &{0.491 $\pm$ 0.015} &{0.569 $\pm$ 0.014} \\
NT-Multispecies (2.5B) &{0.503 $\pm$ 0.010} &{0.481 $\pm$ 0.020} &{0.593 $\pm$ 0.016} &{0.635 $\pm$ 0.016} &{0.481 $\pm$ 0.012} &{0.552 $\pm$ 0.022} \\
\cmidrule(lr){1-7}
Enformer &{0.522 $\pm$ 0.019} &{0.520 $\pm$ 0.015} &{0.552 $\pm$ 0.007} &{0.567 $\pm$ 0.017} &{0.504 $\pm$ 0.021} &{0.626 $\pm$ 0.015} \\
SPACE                    &\textbf{0.548 $\pm$ 0.005} &
\textbf{0.547 $\pm$ 0.007} &{0.586 $\pm$ 0.010} &{0.602 $\pm$ 0.005} &\textbf{0.543 $\pm$ 0.009} &\textbf{0.640 $\pm$ 0.007}\\
\cmidrule(lr){1-7}

\cmidrule(lr){1-7}
\multirow{2}{*}{Model} & 
\multicolumn{4}{c}{\textbf{Chromatin profiles}}&
\multicolumn{2}{c}{\textbf{Regulatory elements}}\\
\cmidrule(lr){2-7}
& {H3K4me3} & {H3K9ac} & {H3K9me3} & {H4K20me1} & {Enhancers} & {Enhancers(types)} \\
\cmidrule(lr){1-7}
DNABERT-2 &{0.646 $\pm$ 0.008} &{0.564 $\pm$ 0.013} &{0.443 $\pm$ 0.025} &{0.655 $\pm$ 0.011} &{0.517 $\pm$ 0.011} &{0.476 $\pm$ 0.009} \\
HyenaDNA-32KB &{0.603 $\pm$ 0.020} &{0.487 $\pm$ 0.025} &{0.419 $\pm$ 0.030} &{0.590 $\pm$ 0.007} &{0.476 $\pm$ 0.021} &{0.445 $\pm$ 0.009} \\
NT-HumanRef (500M) &{0.622 $\pm$ 0.013} &{0.524 $\pm$ 0.013} &{0.433 $\pm$ 0.009} &{0.634 $\pm$ 0.013} &{0.515 $\pm$ 0.019} &{0.477 $\pm$ 0.014} \\
NT-1000G (500M) &{0.609 $\pm$ 0.011} &{0.515 $\pm$ 0.018} &{0.415 $\pm$ 0.019} &{0.634 $\pm$ 0.010} &{0.505 $\pm$ 0.009} &{0.459 $\pm$ 0.011} \\
NT-1000G (2.5B) &{0.615 $\pm$ 0.017} &{0.529 $\pm$ 0.012} &{0.483 $\pm$ 0.013} &\textbf{0.659 $\pm$ 0.008} &{0.504 $\pm$ 0.009} &{0.469 $\pm$ 0.005} \\
NT-Multispecies (2.5B) &{0.618 $\pm$ 0.015} &{0.527 $\pm$ 0.017} &{0.447 $\pm$ 0.018} &{0.650 $\pm$ 0.014} &{0.527 $\pm$ 0.012} &{0.484 $\pm$ 0.012} \\

\cmidrule(lr){1-7}
Enformer &{0.635 $\pm$ 0.019} &{0.593 $\pm$ 0.020} &{0.453 $\pm$ 0.016} &{0.606 $\pm$ 0.016} &{0.614 $\pm$ 0.010} &{0.573 $\pm$ 0.013} \\
SPACE                    &\textbf{0.661 $\pm$ 0.025} &\textbf{0.635 $\pm$ 0.016} &\textbf{0.490 $\pm$ 0.011} &{0.650 $\pm$ 0.011} &\textbf{0.631 $\pm$ 0.007} &\textbf{0.583 $\pm$ 0.008}\\
\cmidrule(lr){1-7}

\cmidrule(lr){1-7}
\multirow{2}{*}{Model} & 
\multicolumn{3}{c}{\textbf{Regulatory elements}}&
\multicolumn{3}{c}{\textbf{Splicing}}\\
\cmidrule(lr){2-7}
&{All} & {NoTATA} & {TATA} & {Donors} & {Acceptors} & {All} \\
\cmidrule(lr){1-7}
DNABERT-2 &{0.754 $\pm$ 0.009} &{0.769 $\pm$ 0.009} &{0.784 $\pm$ 0.036} &{0.837 $\pm$ 0.006} &{0.855 $\pm$ 0.005} &{0.861 $\pm$ 0.004} \\
HyenaDNA-32KB &{0.698 $\pm$ 0.011} &{0.729 $\pm$ 0.009} &{0.666 $\pm$ 0.041} &{0.808 $\pm$ 0.009} &{0.907 $\pm$ 0.018} &{0.915 $\pm$ 0.047} \\
NT-HumanRef (500M) &{0.734 $\pm$ 0.013} &{0.738 $\pm$ 0.008} &{0.831 $\pm$ 0.022} &{0.941 $\pm$ 0.004} &{0.939 $\pm$ 0.003} &{0.952 $\pm$ 0.003} \\
NT-1000G (500M) &{0.727 $\pm$ 0.004} &{0.743 $\pm$ 0.012} &{0.855 $\pm$ 0.041} &{0.933 $\pm$ 0.007} &{0.939 $\pm$ 0.004} &{0.952 $\pm$ 0.004} \\
NT-1000G (2.5B) &{0.708 $\pm$ 0.008} &{0.758 $\pm$ 0.007} &{0.802 $\pm$ 0.030} &{0.952 $\pm$ 0.004} &{0.956 $\pm$ 0.004} &{0.963 $\pm$ 0.001} \\
NT-Multispecies (2.5B) &{0.761 $\pm$ 0.009} &{0.773 $\pm$ 0.010} &\textbf{0.944 $\pm$ 0.016} &\textbf{0.958 $\pm$ 0.003} &\textbf{0.964 $\pm$ 0.003} &\textbf{0.970 $\pm$ 0.002} \\

\cmidrule(lr){1-7}
Enformer &{0.745 $\pm$ 0.012} &{0.763 $\pm$ 0.012} &{0.793 $\pm$ 0.026} &{0.749 $\pm$ 0.007} &{0.739 $\pm$ 0.011} &{0.780 $\pm$ 0.007} \\
SPACE                    &\textbf{0.764 $\pm$ 0.012} &\textbf{0.776 $\pm$ 0.011} &{0.838 $\pm$ 0.028} &{0.942 $\pm$ 0.006} &{0.902 $\pm$ 0.004} &{0.906 $\pm$ 0.003}\\
\cmidrule(lr){1-7}
\bottomrule
\end{tabular}}
\vskip -0.1in
\end{table*}

\subsection{Nucleotide Transformer Downstream Tasks}

We conducted rigorous benchmarking against the suite of 18 genomic datasets established in NT~\citep{NT}, encompassing three fundamental task categories: (1) histone modification marker prediction, (2) cis-regulatory element annotation, and (3) splice site recognition. Following the evaluation protocol from NT, we employed the Matthews Correlation Coefficient (MCC) as the primary performance metric across all tasks to ensure methodological consistency. The formal definition of MCC, along with its theoretical properties, is comprehensively detailed in \cref{sec:mcc}. Our comparative analysis includes both unsupervised pre-training approaches (DNABERT~\citep{DNABert}, DNABERT2~\citep{DNABert2}, and NT~\citep{NT}) and supervised baseline (Enformer~\citep{enformer}). In alignment with NT's methodology, we implemented 10-fold cross-validation with fixed random seeds (0-9) and early stopping based on validation performance. All benchmark performance metrics for the compared models in downstream tasks are directly sourced from the original experimental results reported in NT, ensuring consistent evaluation protocols and dataset configurations. As detailed in Table~\ref{tab:NT_small}, our model achieves SOTA performance on 11 out of 18 prediction tasks. Notably, this superior performance persists even when compared to the parameter-intensive NT-Multispecies variant (2.5B parameters), demonstrating that our supervised pre-training paradigm enables the acquisition of more robust DNA sequence representations. Moreover, our architectural improvements consistently outperform Enformer's original implementation across all tasks, empirically confirming the effectiveness of our modules. The specific details and complete results of the tasks are presented in \cref{appendix:NT}.

% \begin{table*}[htbp]
% \caption{GUE}
% \vskip 0.15in
% \begin{center}
% \label{tab:GUE comparison}
% \resizebox{\textwidth}{!}{ % 缩放表格至页面宽度
% \begin{tabular}{l*{11}{S[table-format=2.2]}}
% \toprule

% \cmidrule(lr){1-6}
% \multirow{2}{*}{Model} & 
% \multicolumn{5}{c}{\textbf{Epigenetic Marks Prediction}}\\
% \cmidrule(lr){2-6}
% & {H3} & {H3K14ac} & {H3K36me3} & {H3K4me1} & {H3K4me2} \\ 
% \cmidrule(lr){1-6}
% Enformer               & {70.65} & {37.87} & {42.41}  & {34.00}  & {29.65}  \\
% SPACE                   & {79.53 ($\uparrow 8.88$)}  & {54.12 ($\uparrow 16.25$)} & {54.82 ($\uparrow 12.41$)} & {50.92($\uparrow 16.92$)} & {43.80 ($\uparrow 14.15$)} \\
% \cmidrule(lr){1-6}

% \cmidrule(lr){1-7}
% \multirow{2}{*}{Model} & 
% \multicolumn{5}{c}{\textbf{Epigenetic Marks Prediction}}& 
% \multicolumn{1}{c}{\textbf{Virus}}\\
% \cmidrule(lr){2-7}
% & {H3K4me3} & {H3K79me3} & {H3K9ac} & {H4} & {H4ac} &{Covid} \\ 
% \cmidrule(lr){2-7}
% Enformer & {22.19}  & {55.69} & {49.35} & {76.32} & {32.90} & {61.33} \\
% SPACE & {49.47 ($\uparrow 27.28$)} & {66.93 ($\uparrow 11.24$)} & {59.29 ($\uparrow 9.94$)} & {81.25 ($\uparrow 4.93$)} & {53.09 ($\uparrow 20.19$)} & {70.26 ($\uparrow 8.93$)} \\
% \cmidrule(lr){1-7}

% \bottomrule
% \end{tabular}}
% \end{center}
% \vskip -0.1in
% \end{table*}

\begin{table*}[htbp]
\caption{\textbf{Comparison Results with Enformer on the GUE Benchmark.} MCC was used in EMP tasks, while F1-score was employed in the CVC task. The upward arrow ($\uparrow$) denotes SPACE's relative improvement over Enformer in novel species.}
\vskip 0.15in
\begin{center}
\label{tab:GUE comparison}
% 第一个表格用更小的字体并居中
\resizebox{0.80\textwidth}{!}{
\begin{tabular}{l*{5}{S[table-format=2.2]}}
\toprule
\multirow{2}{*}{\small Model} & 
\multicolumn{5}{c}{\small\textbf{Epigenetic Marks Prediction}}\\
\cmidrule(lr){2-6}
& {\small H3} & {\small H3K14ac} & {\small H3K36me3} & {\small H3K4me1} & {\small H3K4me2} \\ 
\midrule
{\small Enformer} & {\small 70.65} & {\small 37.87} & {\small 42.41}  & {\small 34.00}  & {\small 29.65}  \\
{\small SPACE} & {\small 79.53 ($\uparrow 8.88$)}  & {\small 54.12 ($\uparrow 16.25$)} & {\small 54.82 ($\uparrow 12.41$)} & {\small 50.92($\uparrow 16.92$)} & {\small 43.80 ($\uparrow 14.15$)} \\
\midrule
\end{tabular}}

\resizebox{0.85\textwidth}{!}{
\begin{tabular}{l*{6}{S[table-format=2.2]}}
\midrule
\multirow{2}{*}{Model} & 
\multicolumn{5}{c}{\textbf{Epigenetic Marks Prediction}}& 
\multicolumn{1}{c}{\textbf{Virus}}\\
\cmidrule(lr){2-6} \cmidrule(lr){7-7}
& {H3K4me3} & {H3K79me3} & {H3K9ac} & {H4} & {H4ac} &{Covid} \\ 
\midrule
Enformer & {22.19}  & {55.69} & {49.35} & {76.32} & {32.90} & {61.33} \\
SPACE & {49.47 ($\uparrow 27.28$)} & {66.93 ($\uparrow 11.24$)} & {59.29 ($\uparrow 9.94$)} & {81.25 ($\uparrow 4.93$)} & {53.09 ($\uparrow 20.19$)} & {70.26 ($\uparrow 8.93$)} \\

\bottomrule
\end{tabular}}

\end{center}
\vskip -0.1in
\end{table*}

\subsection{Cross-species Validation on GUE Benchmark}

To rigorously evaluate the cross-species generalization capacity of our architectural refinements to Enformer, we employed the GUE benchmark~\citep{DNABert2}. While the benchmark encompasses 7 prediction tasks across 4 taxonomic groups, our experimental design strategically prioritizes yeast and viral genomic contexts — evolutionarily distant lineages characterized by marked nucleotide-level divergence from the mammalian species used during model training. These phylogenetically distinct evaluations include Epigenetic Mark Prediction (EMP) across 10 yeast datasets and COVID Variants Classification (CVC) in viral genomes. Following the evaluation protocol established in DNABERT2~\citep{DNABert2}, we adopted the MCC for EMP and the F1-score for CVC. 

As quantified in \cref{tab:GUE comparison}, our refined architecture demonstrates significant improvements over the original Enformer in these tasks. This systematic evaluation provides empirical evidence that our architectural modifications enhance cross-species generalization capabilities, particularly in identifying evolutionarily conserved regulatory features compared to Enformer's baseline implementation. Cross-architecture benchmarking against DNABERT2 and other established baselines (\cref{appendix:GUE}) confirms the universality of these improvements, with non-Enformer baseline results rigorously reproduced from DNABERT2's original experimental protocol to ensure methodological consistency. All evaluations strictly adhered to benchmark specifications, including standardized train-test splits and hyperparameter configurations, to maintain reproducibility and fairness.

\subsection{Genomic Benchmarks}

To further validate the capabilities of our model, we performed extended benchmarking using the Genomic Benchmarks ~\citep{genomic_benchmarks} dataset, which represents the only mainstream benchmark encompassing species beyond those investigated in our previous experiments, including Human-or-worm classification and Drosophila enhancer classification. Following a methodology similar to Caduceus ~\citep{Caduceus}, we evaluated Enformer and SPACE, adopting the baseline model results reported in that paper. It is worth noting that Caduceus did not measure the enhancer prediction task for Drosophila melanogaster, so we referenced the CNN results from Genomic Benchmarks. The results are presented in the ~\cref{tab:genomic}.

\begin{table*}[htbp]
\centering  
\caption{\textbf{The results on the Genomic Benchmarks datasets.} The results for Drosophila enhancers were obtained from Genomic Benchmarks~\citep{genomic_benchmarks}, while all other baseline results were sourced from the Caduceus model~\citep{Caduceus}.}
\vskip 0.15in
\label{tab:genomic}
\resizebox{0.75\textwidth}{!}{
\begin{tabular}{l*{5}{S[table-format=2.2]}}  
\toprule

\cmidrule(lr){1-5}
\multirow{2}{*}{Model} & 
\multicolumn{1}{c}{\textbf{Mouse}}&
\multicolumn{2}{c}{\textbf{Demo}}&
\multicolumn{1}{c}{\textbf{drosophila}}\\
\cmidrule(lr){2-2}\cmidrule(lr){3-4}\cmidrule(lr){5-5}
&\textbf{Enhancers} &\textbf{Coding VS. Intergenomic} &\textbf{Human VS. Worm} &\textbf{Enhancers}\\
\cmidrule(lr){1-5}
CNN         &{0.715 $\pm$ 0.087} &{0.892 $\pm$ 0.008} &{0.942 $\pm$ 0.002} & 0.586 \\
HyenaDNA    &{0.780 $\pm$ 0.025} &{0.904 $\pm$ 0.005} &{0.964 $\pm$ 0.002} & {$-$} \\
Mamba       &{0.743 $\pm$ 0.054} &{0.904 $\pm$ 0.004} &{0.967 $\pm$ 0.002} & {$-$} \\
Caduceus-PH &{0.754 $\pm$ 0.074} &{0.915 $\pm$ 0.003} &\textbf{0.973 $\pm$ 0.001} & {$-$} \\
Caduceus-PS &{0.793 $\pm$ 0.058} &{0.910 $\pm$ 0.003} &{0.968 $\pm$ 0.002} & {$-$} \\
\cmidrule(lr){1-5}
Enformer    &{0.835 $\pm$ 0.012} &{0.913 $\pm$ 0.001} &{0.958 $\pm$ 0.001} &{0.613 $\pm$ 0.005} \\
SPACE       &\textbf{0.905 $\pm$ 0.010} &\textbf{0.922 $\pm$ 0.001} &{0.967 $\pm$ 0.004} &\textbf{0.721 $\pm$ 0.016} \\
\cmidrule(lr){1-5}
\end{tabular}}

\resizebox{0.75\textwidth}{!}{ 
\begin{tabular}{l*{6}{S[table-format=2.2]}}  
\cmidrule(lr){1-6}
\multirow{2}{*}{Model} &
\multicolumn{5}{c}{\textbf{Human}}\\
\cmidrule(lr){2-6}
& \textbf{Enhancers Cohn} & \textbf{Enhancer Ensembl} & \textbf{Regulatory} & \textbf{OCR Ensembl} & \textbf{Nontata Promoters}\\
\cmidrule(lr){1-6}
CNN         &{0.702 $\pm$ 0.021} &{0.744 $\pm$ 0.122} &{0.872 $\pm$ 0.005} &{0.698 $\pm$ 0.013} &{0.861 $\pm$ 0.009} \\
HyenaDNA    &{0.729 $\pm$ 0.014} &{0.849 $\pm$ 0.006} &{0.869 $\pm$ 0.012} &{0.783 $\pm$ 0.007} &{0.944 $\pm$ 0.002} \\
Mamba       &{0.732 $\pm$ 0.029} &{0.862 $\pm$ 0.008} &{0.814 $\pm$ 0.211} &{0.815 $\pm$ 0.002} &{0.933 $\pm$ 0.007} \\
Caduceus-PH &{0.747 $\pm$ 0.004} &{0.893 $\pm$ 0.008} &{0.872 $\pm$ 0.011} &{0.828 $\pm$ 0.006} &\textbf{0.946 $\pm$ 0.007} \\
Caduceus-PS &{0.745 $\pm$ 0.007} &{0.900 $\pm$ 0.006} &{0.873 $\pm$ 0.007} &{0.818 $\pm$ 0.006} &{0.945 $\pm$ 0.010} \\
\cmidrule(lr){1-6}
Enformer    &{0.723 $\pm$ 0.001} &{0.844 $\pm$ 0.001} &{0.903 $\pm$ 0.001} &\textbf{0.876 $\pm$ 0.001} &{0.878 $\pm$ 0.002} \\ 
SPACE       &\textbf{0.769 $\pm$ 0.006} &\textbf{0.919 $\pm$ 0.014} &\textbf{0.944 $\pm$ 0.002} &{0.854 $\pm$ 0.001} &{0.940 $\pm$ 0.002} \\ 
\cmidrule(lr){1-6}

\bottomrule
\end{tabular}}
\vskip -0.1in
\end{table*}

\begin{figure*}[ht]
\vskip 0.2in
 \centering
 \begin{minipage}[c]{0.25\textwidth} 
   \begin{subfigure}[b]{\textwidth}
     \includegraphics[width=\textwidth]{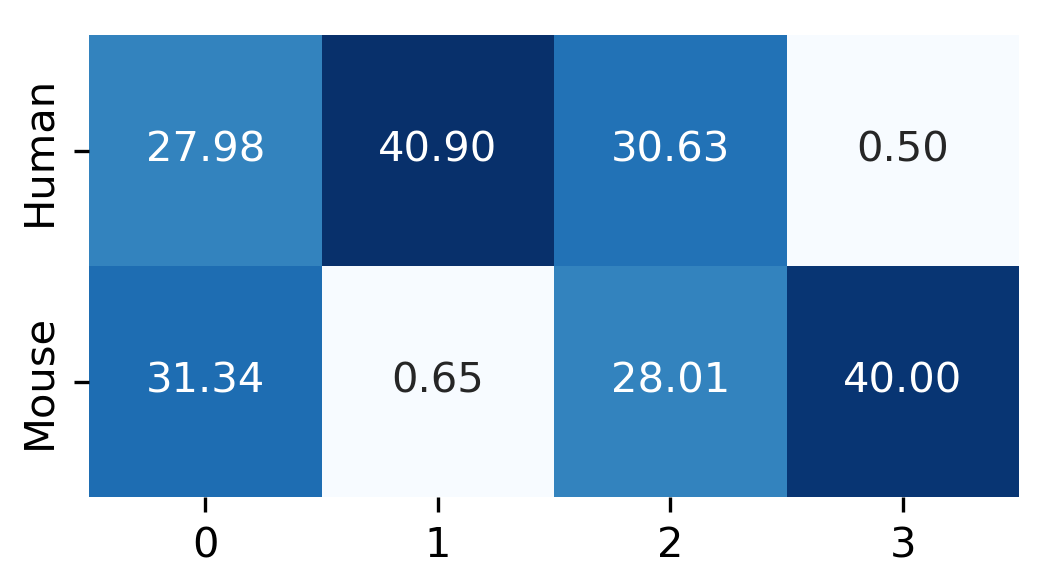}
     \caption{}
     \label{species_gate}
   \end{subfigure}
   
   \vspace{0.3cm}
   
   \begin{subfigure}[b]{\textwidth}
     \includegraphics[width=\textwidth]{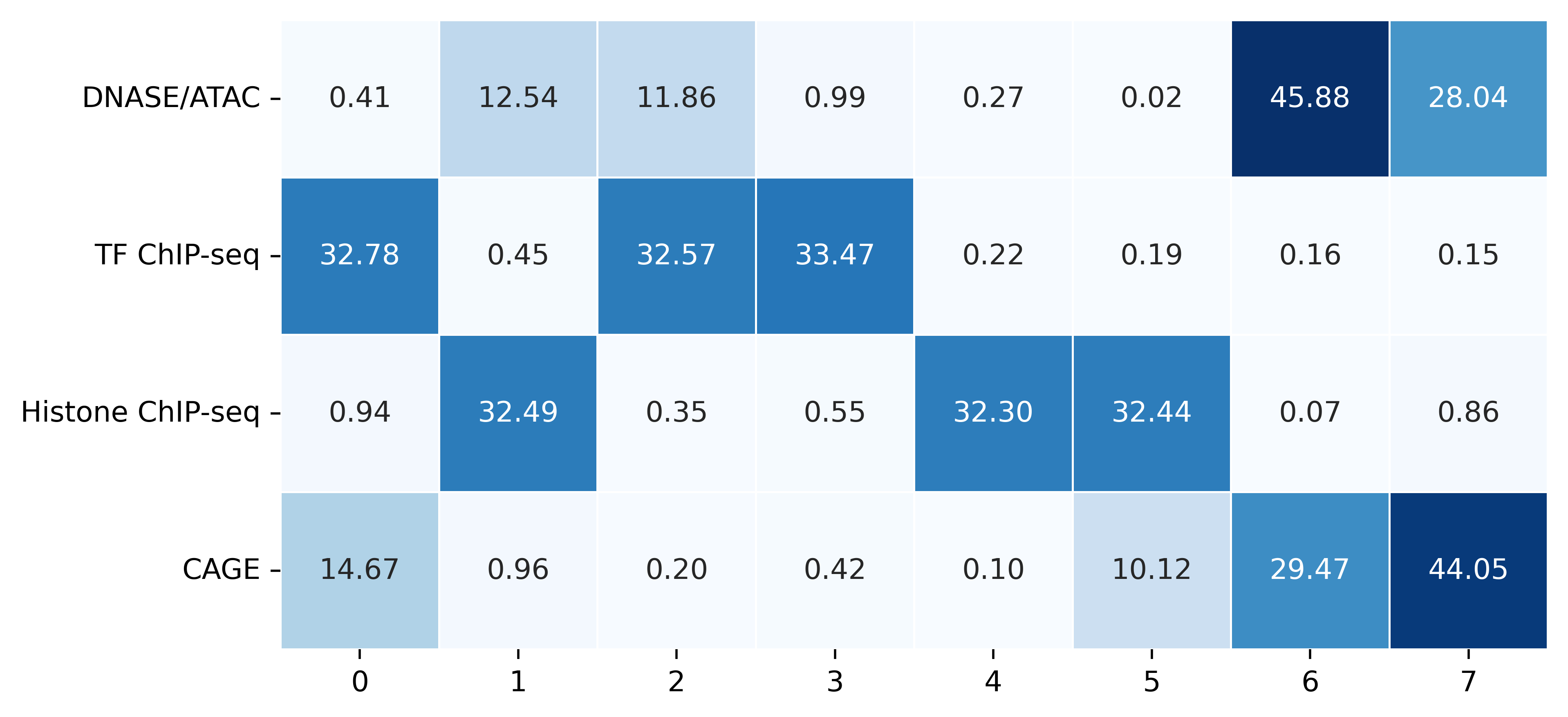}
     \caption{}
     \label{profiles_gate}
   \end{subfigure}
 \end{minipage}
 \hspace{0.02\textwidth}
 \begin{minipage}[c]{0.70\textwidth} 
   \begin{subfigure}[b]{\textwidth}
     \includegraphics[width=\textwidth]{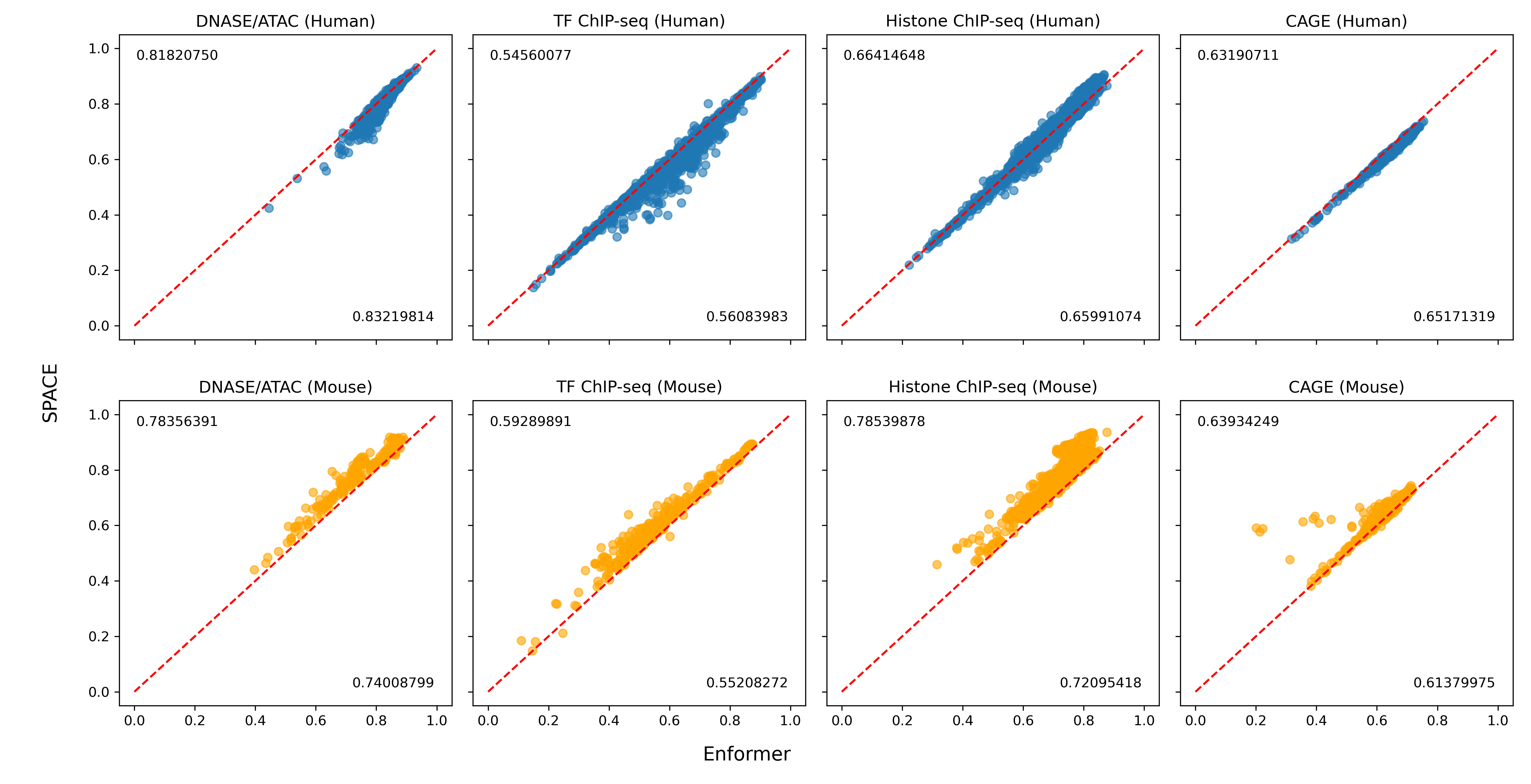}
     \caption{}
     \label{prediction_results}
   \end{subfigure}
 \end{minipage}
 
 \caption{
   \textbf{Expert selection visualizations and prediction results.}
   (\subref{species_gate}) Visualization of expert selection in the final cross-species MoE.
   (\subref{profiles_gate}) Expert selection in the profile-grouped enhancement decoder module.
   (\subref{prediction_results}) Pearson correlation coefficients across all positions per profile on the test set. Each point represents the average correlation of predicted genomic profiles across all 128-bp binned genomic positions.
 }
 \label{merged_figures}
\vskip -0.2in
\end{figure*}

\subsection{Analysis of the MoE architecture}

\textbf{Analysis of the species-aware encoder module.}
As shown in \cref{species_gate}, this part reveals the implicit species-characteristic learning mechanism through visual analysis of expert selection frequencies in the final Transformer layer from the Enformer test dataset. In our experiments, we adopt a 4-expert architecture with a top-3 selection mechanism ($k=3$). Quantitative analysis demonstrates significant functional differentiation in the expert system: Expert 1 and Expert 3 exhibit species-specific learning capabilities for single species (human and mouse), while Expert 0 and Expert 2 primarily participate in cross-species conserved feature extraction. This hierarchical division successfully achieves spatial decoupling of species-specific representations and evolutionarily conserved features, providing an interpretable solution for multi-species joint modeling.

\textbf{Analysis of the profile-grouped enhancement decoder module.}
In our experiments, we employed 8 cross-profile-type shared experts (\(K=8\)) with 2 expert-selected groups per profile type (\(R=2\)), where each group dynamically integrated the top 3 most contributory experts through a dual-gated expert weighted aggregation. Through hierarchical weighting, we derived final expert selection probabilities. Normalized expert selection frequencies across Enformer test datasets were systematically quantified and visualized in \cref{profiles_gate}, revealing distinct biological patterns.  

Notably, TF binding (TF ChIP-seq) and histone modifications (Histone ChIP-seq) profiles exhibited high expert specialization, reflecting their biologically inherent complexity: TF binding involves combinatorial interactions among diverse transcription factor families, while histone modifications require interpretation of multilayered epigenetic codes governed by cooperative post-translational modifications. Conversely, Chromatin accessibility (DNase/ATAC-seq signals) and transcription initiation (CAGE signals) profiles showed substantial expert overlap with differential weighting, mirroring their mechanistic and positional interdependence. Chromatin accessibility creates permissive 3D topological environments essential for transcription initiation, with transcription start site (TSS)-associated promoter and enhancer regions exhibiting substantial spatial overlap with chromatin accessibility domains. This ``functional dependency-spatial coupling'' drives the prediction enhancement decoder module to develop coordinated feature extraction strategies. These findings collectively demonstrate that divergent genomic profiles converge on shared regulatory architectures and engage in reciprocal regulatory interactions – a phenomenon that fundamentally informed our development of the profile-grouped enhancement decoder to systematically leverage such interconnectivity.

\subsection{Comparative Analysis with Enformer in Gene Expression Prediction}

We conducted a comparative analysis based on the core task of the baseline Enformer model, which aims to predict human and mouse genomic profiles at 128-bp resolution from input DNA sequences. We computed the average Pearson correlation coefficients across all positions for genomic profiles in the test set and performed stratified visualization by species and profile types, as illustrated in \cref{prediction_results}. The results demonstrate that our approach significantly enhances the prediction accuracy for mouse genomic profiles while maintaining the prediction performance for human genomic profiles. It should be noted that our training steps were only 1/3 of those used by Enformer, yet we were still able to achieve good results on the training objective tasks.

\subsection{Ablation Study}

We conducted ablation experiments with a half-scale model (hidden\_dim=768) on five configurations: (1) baseline without the prediction-enhanced decoder, (2) decoder replacement with a parameter-matched MLP, (3) substitution of MoE layers with standard FFNs in the encoder, (4) additional removal of species embeddings from configuration (3), and (5) our complete dual-module architecture. The results are presented in the \cref{tab:ablation}. SPACE demonstrates superior performance across most tasks, with the notable exception of the TATA box dataset (see~\cref{tab:ablation_NT}) -- due to its exclusive focus on simple sequence motifs rather than complex regulatory mechanisms. This indicates that while our decoder doesn't directly boost chromatin profile prediction accuracy, the MoE architecture implicitly models cross-profile regulatory dependencies, offering significant advantages for tasks requiring integrated profile understanding. Cross-species evaluation on the GUE benchmark (yeast and virus tasks, detailed in~\cref{tab:ablation_GUE}) further demonstrates that the MLP-based decoder variant exhibits substantially weaker generalization to new species compared to SPACE's enhancement decoder architecture.

\begin{table*}[htbp]
\caption{\textbf{Ablation Studies on NT downstream tasks and GUE benchmarks.} The results include the average outcomes of the three major categories of downstream tasks in NT and the average results of the EMP tasks and the CVC task in the GUE benchmark experiments.}
\vskip 0.15in
\centering
\label{tab:ablation}
\resizebox{0.8\textwidth}{!}{
\begin{tabular}{l*{6}{S[table-format=1.3]}}
\toprule

\cmidrule(lr){1-6}
\multirow{2}{*}{Model} & 
\multicolumn{3}{c}{\textbf{NT}}&
\multicolumn{2}{c}{\textbf{GUE}} \\
\cmidrule(lr){2-4}\cmidrule{5-6}
& {Chromatin} & {Regulatory} & {Splicing} & {EMP} & {CVC}  \\
\cmidrule(lr){1-6}
SPACE w/o decoder            &0.5674 &0.7054 &0.8977 &0.5339 &0.6866 \\                
SPACE w/o decoder w/ MLP          &0.5651 &0.6920 &0.9020 &0.5153 &0.6783 \\   
SPACE w/o encoder                 &0.5653 &0.7022 &0.8887 &0.5346 &0.6846 \\     
SPACE w/o encoder w/o species emb &0.5692 &0.6986 &0.8957 &0.5322 &0.6856 \\        
SPACE                             &0.5705 &0.7024 &0.9077 &0.5368 &0.6889 \\  
\cmidrule(lr){1-6}

\bottomrule
\end{tabular}}
\vskip -0.1in
\end{table*}

\section{Limitations}
This work has limitations in data coverage and model scale. First, SPACE has only been trained on two species. While this study demonstrates the advantages of our design, we believe extending to more species would yield benefits as additional data becomes available~\citep{vandereyken2023methods}. Second, constrained by computational resources, our model (588M, sparse-activated) is significantly smaller than the largest variant of NT (2.5B, dense) (the detail is in \cref{appendix:parameter}. Given the observed scaling laws in DFMs~\citep{NT, nguyen2024sequence}, we believe increasing our model scale would lead to improvements.

\section{Conclusion}
In this work, we demonstrate that supervised pre-training via genomic profile prediction serves as a moreeffective and targeted alternative to pure sequence pre-training for DNA foundation models. To distinguish unique characteristics of different species and profiles while fully utilizing the transferrable knowledge among species and profiles, we introduce SPACE that provides biological insights into the model architecture. Through extensive evaluations, we show that our supervised pre-training with the proposed SPACE sets a new standard for DNA representation learning, paving the way for future developments in DFMs.

\section*{Impact Statement}
This paper presents work whose goal is to advance the field at the intersection of Machine Learning and Genomics. There are many potential societal consequences of our work in this interdisciplinary area, none of which we feel must be specifically highlighted here.
\section*{Acknowledgements}
This work was supported in part by the National Natural Science Foundation of China No. 62376277 and No. 61976206, the Public Computing Cloud of Renmin University of China, and the Fund for Building World-Class Universities (Disciplines) of Renmin University of China.

\newpage

\bibliography{example_paper}
\bibliographystyle{icml2025}

%%%%%%%%%%%%%%%%%%%%%%%%%%%%%%%%%%%%%%%%%%%%%%%%%%%%%%%%%%%%%%%%%%%%%%%%%%%%%%%
%%%%%%%%%%%%%%%%%%%%%%%%%%%%%%%%%%%%%%%%%%%%%%%%%%%%%%%%%%%%%%%%%%%%%%%%%%%%%%%
% APPENDIX
%%%%%%%%%%%%%%%%%%%%%%%%%%%%%%%%%%%%%%%%%%%%%%%%%%%%%%%%%%%%%%%%%%%%%%%%%%%%%%%
%%%%%%%%%%%%%%%%%%%%%%%%%%%%%%%%%%%%%%%%%%%%%%%%%%%%%%%%%%%%%%%%%%%%%%%%%%%%%%%
\newpage
\appendix
\onecolumn
\section{Derivation of Mathematical Formulations for Key Functions}

\subsection{Mutual Information Analysis}
\label{appendix:MI}

The Mutual Information defined in \cref{eq:MI} is:
\begin{equation*}
\begin{aligned}
\mathcal{L}_{\text{MI}} &= - MI(S;E) = -H(S) - H(E) + H(S,E) \\
&= \sum_{i=m}^M P(S_m)\log P(S_m) + \sum_{n=1}^N P(E_n)\log P(E_n) \\
&\quad - \sum_{m=1}^M\sum_{n=1}^N P(S_m,E_n)\log P(S_m,E_n),
\end{aligned}
\end{equation*}
where $S_m$ denotes the species probability and $E_n$ represents the selection weight of each expert.

We split the formulae to analyse them separately. The mutual information decomposition exhibits three fundamental components:

\textbf{Species Entropy}: 
    \begin{equation*}
    -\sum_{i=1}^M P(S_i)\log P(S_i) = H(S).
    \end{equation*}
    
    This term represents the inherent diversity of species distribution in training data. As $P(S_i)$ constitutes a fixed prior, $H(S)$ remains constant during optimization.

\textbf{Expert Diversity Regularization}:
    \begin{equation*}
    -\sum_{j=1}^N P(E_j)\log P(E_j) = H(E).
    \end{equation*}
    
    Maximizing this entropy term encourages balanced utilization of experts, preventing expert collapse where few experts dominate computations. Formally, this ensures:
    \begin{equation*}
        \lim_{H(E) \to \log N} P(E_j) = \frac{1}{N}\quad \forall j.
    \end{equation*}

\textbf{Conditional Specialization Objective}: 
    \begin{equation*}
    \sum_{i=1}^M\sum_{j=1}^N P(S_i,E_j)\log P(S_i,E_j) = -H(S,E).
    \end{equation*}

    Minimizing this joint entropy (equivalent to maximizing $-H(S,E)$) sharpens the conditional distribution $P(E_j|S_i)$, thereby promoting:
    \begin{equation*}
        \lim_{H(S,E) \to 0} P(E_j|S_i) = \begin{cases} 
        1 & \text{if } j = \arg\max_k G^{S_i}_k(x) \\
        0 & \text{otherwise}
        \end{cases}.
    \end{equation*}
    This objective ensures that, for a given species, the model preferentially activates a fixed subset of $k$ experts.

In this way, the sparse MoE-based encoding module encourages different expert combinations to handle different species, while some shared experts in the pool can capture common knowledge across species.

\subsection{Poisson negative log-likelihood}
\label{sec:poisson}
The Poisson negative log-likelihood function is defined as 

\begin{equation*}
\mathcal{L}_{\text{Poisson}} = \frac{1}{N} \sum_{i=1}^N \left( p_i - t_i \ln p_i \right),
\end{equation*}

whree $p$ denotes the prediction vector and $t$ represents the target vector.

\subsection{Matthews Correlation Coefficient (MCC)}
\label{sec:mcc}
The Matthews Correlation Coefficient (MCC) is a statistically rigorous metric for evaluating classification models. Its definition and generalization to multi-class problems are formally outlined below.  

\textbf{Binary Classification Case}
For binary classification, let \( TP \), \( TN \), \( FP \), and \( FN \) denote the counts of true positives, true negatives, false positives, and false negatives, respectively. The MCC is defined as:  

\[
\text{MCC} = \frac{TP \cdot TN - FP \cdot FN}{\sqrt{(TP + FP)(TP + FN)(TN + FP)(TN + FN)}}.
\]  

Here, \( TP \), \( TN \), \( FP \), and \( FN \) correspond to entries in the confusion matrix for two classes.  

\textbf{Multi-class Classification Case}

For \( K \)-class classification (\( K \geq 2 \)), let \( C \) be the \( K \times K \) confusion matrix, where \( C_{ij} \) represents the number of samples from class \( i \) predicted as class \( j \). The MCC generalizes to:  

\[
\text{MCC} = \frac{\sum_{k=1}^K \sum_{l=1}^K \sum_{m=1}^K C_{kk} C_{lm} - C_{kl} C_{mk}}{\sqrt{\left(\sum_{k=1}^K \sum_{l=1}^K C_{kl} \sum_{\substack{m=1 \\ m \neq k}}^K C_{mk}\right) \left(\sum_{k=1}^K \sum_{l=1}^K C_{lk} \sum_{\substack{m=1 \\ m \neq k}}^K C_{km}\right)}}.
\]  

This formulation quantifies the covariance between all class pairs, ensuring robustness to imbalanced data distributions.  

The MCC ranges in  \( [-1, 1] \), where \( 1 \), \( 0 \), and \( -1 \) correspond to perfect prediction, random guessing, and total disagreement, respectively.

\section{Pre-training Dataset}
\label{appendix:Pre-training Dataset}

\begin{table}[H]
\centering
\caption{Genomic Dataset Statistics}
\vskip 0.15in
\label{tab:dataset_composition}
\begin{tabular}{lcccc}
\toprule
Species & Train & Val & Test & Sequence Length \\
\midrule
Human & 34,021 & 2,213 & 1,937 & 131,072 bp \\
Mouse & 29,295 & 2,209 & 2,017 & 131,072 bp \\
\bottomrule
\end{tabular}
\vskip -0.1in
\end{table}

Our model was pretrained on the same dataset as Enformer~\citep{enformer}, with detailed composition statistics provided in \cref{tab:dataset_composition}. To address the pronounced species imbalance between human and mouse genomic data, we implemented balanced batch sampling through randomized minority-class augmentation, ensuring equal representation of both species in every batch. This strategy mitigates species bias while preserving sequence diversity through stochastic resampling.

The dataset comprises DNA sequences paired with genomic profiles as prediction targets. These genomic profiles are categorized into four functional classes: chromatin accessibility (DNase/ATAC-seq), transcription factor binding (TF ChIP-seq), histone modifications (Histone ChIP-seq), and transcriptional activity (CAGE). The species-specific distribution of profile types is quantified in \cref{tab:profiles_num}, which details the number of available tracks per category for each organism.

\begin{table}[h]
\centering
\caption{Distribution of Genomics profiles }
\vskip 0.15in
\label{tab:profiles_num}

\begin{tabular}{lccccc}
\toprule
species & DNase/ATA & TF ChIP & Histone ChIP & CAGE & Total\\
\midrule
Human  &684 &2131 &1860 &638 &5313 \\
Mouse &228 &308 &750 &357 &1643 \\
\bottomrule
\end{tabular}
\vskip -0.1in
\end{table}

\section{Nucleotide Transformer Downstream Tasks Revised}
\label{appendix:NT}

\subsection{Datasets}

The benchmark dataset comprises 18 downstream tasks originally proposed in Nucleotide Transformer (NT) \citep{NT}, accessible via \url{https://huggingface.co/datasets/InstaDeepAI/nucleotide_transformer_downstream_tasks_revised}. These tasks establish a unified genomics benchmarking framework encompassing both binary and multi-class classification challenges. All data is exclusively derived from human samples, organized into three biologically meaningful categories: Chromatin Profiles, Regulatory Elements and Splicing. The complete dataset composition, including sequence numbers, class distributions and sequence length statistics, is detailed in \cref{tab:NT Dataset}.

\begin{table}[htbp]
\centering
\caption{ Details of the NT downstream tasks}
\vskip 0.15in
\label{tab:NT Dataset}
\begin{tabular}{lcccc}
\toprule
\textbf{Task} & \textbf{Number of train sequences} & \textbf{Number of test sequences} & \textbf{Number of labels} & \textbf{Sequence length} \\ 
\midrule
promoter\_all & 30,000 & 1,584 & 2 & 300 \\ 
promoter\_tata & 5,062 & 212 & 2 & 300 \\ 
promoter\_no\_tata & 30,000 & 1,372 & 2 & 300 \\ 
enhancers & 30,000 & 3,000 & 2 & 400 \\ 
enhancers\_types & 30,000 & 3,000 & 3 & 400 \\ 
splice\_sites\_all & 30,000 & 3,000 & 3 & 600 \\ 
splice\_sites\_acceptor & 30,000 & 3,000 & 2 & 600 \\ 
splice\_sites\_donor & 30,000 & 3,000 & 2 & 600 \\ 
H2AFZ & 30,000 & 3,000 & 2 & 1,000 \\ 
H3K27ac & 30,000 & 1,616 & 2 & 1,000 \\ 
H3K27me3 & 30,000 & 3,000 & 2 & 1,000 \\ 
H3K36me3 & 30,000 & 3,000 & 2 & 1,000 \\ 
H3K4me1 & 30,000 & 3,000 & 2 & 1,000 \\ 
H3K4me2 & 30,000 & 2,138 & 2 & 1,000 \\ 
H3K4me3 & 30,000 & 776 & 2 & 1,000 \\ 
H3K9ac & 23,274 & 1,004 & 2 & 1,000 \\ 
H3K9me3 & 27,438 & 850 & 2 & 1,000 \\ 
H4K20me1 & 30,000 & 2,270 & 2 & 1,000 \\ 
\bottomrule
\end{tabular}
\vskip -0.1in
\end{table}

\subsection{Implementation}

We maintained identical hyperparameter configurations across all tasks. Our systematic hyperparameter search included learning rates of $5 \times 10^{-5}$, $3 \times 10^{-5}$, and $5 \times 10^{-4}$, combined with batch sizes of 8, 16, and 32. Through empirical validation, we identified the optimal configuration employing a learning rate of $5 \times 10^{-5}$ with batch size 8. The training protocol utilized the AdamW optimizer \citep{adamw} over 3 epochs, while retaining default parameter settings from the HuggingFace Transformer Trainer implementation \citep{wolf-etal-2020-transformers}.

\subsection{Results}

The complete benchmark results of the downstream tasks for NT are presented in \cref{tab:NT}. All baseline results are sourced from NT \citep{NT}. Performance per task was calculated as the median of the 10 cross-validation folds ($\pm$ standard deviation). The best results for each task are highlighted in \textbf{bold}. 

\begin{table}[H]
\caption{Complete Benchmark Results of Nucleotide Transformer 
 Downstream Tasks}
\vskip 0.15in
\centering
\label{tab:NT}
\resizebox{\textwidth}{!}{ % 缩放表格至页面宽度
\begin{tabular}{l*{6}{S[table-format=2.2]}}
\toprule

\cmidrule(lr){1-7}
\multirow{2}{*}{Model} & 
\multicolumn{6}{c}{\textbf{Chromatin profiles}}\\
\cmidrule(lr){2-7}
& {H2AFZ} & {H3K27ac} & {H3K27me3} & {H3K36me3} & {H3K4me1} & {H3K4me2} \\
\cmidrule(lr){1-7}
BPNet (original) &{0.473 $\pm$ 0.009} &{0.296 $\pm$ 0.046} &{0.543 $\pm$ 0.009} &{0.548 $\pm$ 0.009} &{0.436 $\pm$ 0.008} &{0.427 $\pm$ 0.036} \\
BPNet (large) &{0.487 $\pm$ 0.014} &{0.214 $\pm$ 0.037} &{0.551 $\pm$ 0.009} &{0.570 $\pm$ 0.009} &{0.459 $\pm$ 0.012} &{0.427 $\pm$ 0.025} \\
DNABERT-2 &{0.490 $\pm$ 0.013} &{0.491 $\pm$ 0.010} &{0.599 $\pm$ 0.010} &\textbf{0.637 $\pm$ 0.007} &{0.490 $\pm$ 0.008} &{0.558 $\pm$ 0.013} \\
HyenaDNA-1KB &{0.455 $\pm$ 0.015} &{0.423 $\pm$ 0.017} &{0.541 $\pm$ 0.018} &{0.543 $\pm$ 0.010} &{0.430 $\pm$ 0.014} &{0.521 $\pm$ 0.024} \\
HyenaDNA-32KB &{0.467 $\pm$ 0.012} &{0.421 $\pm$ 0.010} &{0.550 $\pm$ 0.009} &{0.553 $\pm$ 0.011} &{0.423 $\pm$ 0.016} &{0.515 $\pm$ 0.018} \\
NT-HumanRef (500M) &{0.465 $\pm$ 0.011} &{0.457 $\pm$ 0.010} &{0.589 $\pm$ 0.009} &{0.594 $\pm$ 0.004} &{0.468 $\pm$ 0.007} &{0.527 $\pm$ 0.011} \\
NT-1000G (500M) &{0.464 $\pm$ 0.012} &{0.458 $\pm$ 0.012} &{0.591 $\pm$ 0.007} &{0.581 $\pm$ 0.009} &{0.466 $\pm$ 0.006} &{0.528 $\pm$ 0.011} \\
NT-1000G (2.5B) &{0.478 $\pm$ 0.012} &{0.486 $\pm$ 0.023} &\textbf{0.603 $\pm$ 0.009} &{0.632 $\pm$ 0.008} &{0.491 $\pm$ 0.015} &{0.569 $\pm$ 0.014} \\
NT-Multispecies (2.5B) &{0.503 $\pm$ 0.010} &{0.481 $\pm$ 0.020} &{0.593 $\pm$ 0.016} &{0.635 $\pm$ 0.016} &{0.481 $\pm$ 0.012} &{0.552 $\pm$ 0.022} \\
\cmidrule(lr){1-7}
Enformer &{0.522 $\pm$ 0.019} &{0.520 $\pm$ 0.015} &{0.552 $\pm$ 0.007} &{0.567 $\pm$ 0.017} &{0.504 $\pm$ 0.021} &{0.626 $\pm$ 0.015} \\
SPACE                    &\textbf{0.548 $\pm$ 0.005} &
\textbf{0.547 $\pm$ 0.007} &{0.586 $\pm$ 0.010} &{0.602 $\pm$ 0.005} &\textbf{0.543 $\pm$ 0.009} &\textbf{0.640 $\pm$ 0.007}\\
\cmidrule(lr){1-7}

\cmidrule(lr){1-7}
\multirow{2}{*}{Model} & 
\multicolumn{4}{c}{\textbf{Chromatin profiles}}&
\multicolumn{2}{c}{\textbf{Regulatory elements}}\\
\cmidrule(lr){2-7}
& {H3K4me3} & {H3K9ac} & {H3K9me3} & {H4K20me1} & {Enhancers} & {Enhancers(types)} \\
\cmidrule(lr){1-7}
BPNet (original) &{0.445 $\pm$ 0.047} &{0.336 $\pm$ 0.034} &{0.298 $\pm$ 0.030} &{0.531 $\pm$ 0.025} &{0.488 $\pm$ 0.009} &{0.449 $\pm$ 0.006} \\
BPNet (large) &{0.445 $\pm$ 0.049} &{0.298 $\pm$ 0.033} &{0.234 $\pm$ 0.037} &{0.525 $\pm$ 0.038} &{0.492 $\pm$ 0.008} &{0.454 $\pm$ 0.008} \\
DNABERT-2 &{0.646 $\pm$ 0.008} &{0.564 $\pm$ 0.013} &{0.443 $\pm$ 0.025} &{0.655 $\pm$ 0.011} &{0.517 $\pm$ 0.011} &{0.476 $\pm$ 0.009} \\
HyenaDNA-1KB &{0.596 $\pm$ 0.015} &{0.484 $\pm$ 0.022} &{0.375 $\pm$ 0.026} &{0.580 $\pm$ 0.009} &{0.475 $\pm$ 0.006} &{0.441 $\pm$ 0.010} \\
HyenaDNA-32KB &{0.603 $\pm$ 0.020} &{0.487 $\pm$ 0.025} &{0.419 $\pm$ 0.030} &{0.590 $\pm$ 0.007} &{0.476 $\pm$ 0.021} &{0.445 $\pm$ 0.009} \\
NT-HumanRef (500M) &{0.622 $\pm$ 0.013} &{0.524 $\pm$ 0.013} &{0.433 $\pm$ 0.009} &{0.634 $\pm$ 0.013} &{0.515 $\pm$ 0.019} &{0.477 $\pm$ 0.014} \\
NT-1000G (500M) &{0.609 $\pm$ 0.011} &{0.515 $\pm$ 0.018} &{0.415 $\pm$ 0.019} &{0.634 $\pm$ 0.010} &{0.505 $\pm$ 0.009} &{0.459 $\pm$ 0.011} \\
NT-1000G (2.5B) &{0.615 $\pm$ 0.017} &{0.529 $\pm$ 0.012} &{0.483 $\pm$ 0.013} &\textbf{0.659 $\pm$ 0.008} &{0.504 $\pm$ 0.009} &{0.469 $\pm$ 0.005} \\
NT-Multispecies (2.5B) &{0.618 $\pm$ 0.015} &{0.527 $\pm$ 0.017} &{0.447 $\pm$ 0.018} &{0.650 $\pm$ 0.014} &{0.527 $\pm$ 0.012} &{0.484 $\pm$ 0.012} \\

\cmidrule(lr){1-7}
Enformer &{0.635 $\pm$ 0.019} &{0.593 $\pm$ 0.020} &{0.453 $\pm$ 0.016} &{0.606 $\pm$ 0.016} &{0.614 $\pm$ 0.010} &{0.573 $\pm$ 0.013} \\
SPACE                    &\textbf{0.661 $\pm$ 0.025} &\textbf{0.635 $\pm$ 0.016} &\textbf{0.490 $\pm$ 0.011} &{0.650 $\pm$ 0.011} &\textbf{0.631 $\pm$ 0.007} &\textbf{0.583 $\pm$ 0.008}\\
\cmidrule(lr){1-7}

\cmidrule(lr){1-7}
\multirow{2}{*}{Model} & 
\multicolumn{3}{c}{\textbf{Regulatory elements}}&
\multicolumn{3}{c}{\textbf{Splicing}}\\
\cmidrule(lr){2-7}
&{All} & {NoTATA} & {TATA} & {Donors} & {Acceptors} & {All} \\
\cmidrule(lr){1-7}
BPNet (original) &{0.696 $\pm$ 0.026} &{0.717 $\pm$ 0.023} &{0.848 $\pm$ 0.042} &{0.859 $\pm$ 0.038} &{0.793 $\pm$ 0.072} &{0.920 $\pm$ 0.014} \\
BPNet (large) &{0.672 $\pm$ 0.023} &{0.672 $\pm$ 0.043} &{0.826 $\pm$ 0.017} &{0.925 $\pm$ 0.031} &{0.865 $\pm$ 0.026} &{0.930 $\pm$ 0.021} \\
DNABERT-2 &{0.754 $\pm$ 0.009} &{0.769 $\pm$ 0.009} &{0.784 $\pm$ 0.036} &{0.837 $\pm$ 0.006} &{0.855 $\pm$ 0.005} &{0.861 $\pm$ 0.004} \\
HyenaDNA-1KB &{0.693 $\pm$ 0.016} &{0.723 $\pm$ 0.013} &{0.648 $\pm$ 0.044} &{0.815 $\pm$ 0.049} &{0.854 $\pm$ 0.053} &{0.943 $\pm$ 0.024} \\
HyenaDNA-32KB &{0.698 $\pm$ 0.011} &{0.729 $\pm$ 0.009} &{0.666 $\pm$ 0.041} &{0.808 $\pm$ 0.009} &{0.907 $\pm$ 0.018} &{0.915 $\pm$ 0.047} \\
NT-HumanRef (500M) &{0.734 $\pm$ 0.013} &{0.738 $\pm$ 0.008} &{0.831 $\pm$ 0.022} &{0.941 $\pm$ 0.004} &{0.939 $\pm$ 0.003} &{0.952 $\pm$ 0.003} \\
NT-1000G (500M) &{0.727 $\pm$ 0.004} &{0.743 $\pm$ 0.012} &{0.855 $\pm$ 0.041} &{0.933 $\pm$ 0.007} &{0.939 $\pm$ 0.004} &{0.952 $\pm$ 0.004} \\
NT-1000G (2.5B) &{0.708 $\pm$ 0.008} &{0.758 $\pm$ 0.007} &{0.802 $\pm$ 0.030} &{0.952 $\pm$ 0.004} &{0.956 $\pm$ 0.004} &{0.963 $\pm$ 0.001} \\
NT-Multispecies (2.5B) &{0.761 $\pm$ 0.009} &{0.773 $\pm$ 0.010} &\textbf{0.944 $\pm$ 0.016} &\textbf{0.958 $\pm$ 0.003} &\textbf{0.964 $\pm$ 0.003} &\textbf{0.970 $\pm$ 0.002} \\

\cmidrule(lr){1-7}
Enformer &{0.745 $\pm$ 0.012} &{0.763 $\pm$ 0.012} &{0.793 $\pm$ 0.026} &{0.749 $\pm$ 0.007} &{0.739 $\pm$ 0.011} &{0.780 $\pm$ 0.007} \\
SPACE                    &\textbf{0.764 $\pm$ 0.012} &\textbf{0.776 $\pm$ 0.011} &{0.838 $\pm$ 0.028} &{0.942 $\pm$ 0.006} &{0.902 $\pm$ 0.004} &{0.906 $\pm$ 0.003}\\
\cmidrule(lr){1-7}
\bottomrule
\end{tabular}}
\vskip -0.1in
\end{table}

\section{GUE}
\label{appendix:GUE}

\subsection{Dataset}

GUE is a comprehensive benchmark for genome understanding consising of 28 distinct datasets across 7 tasks and 4 species, downloaded from \url{https://github.com/MAGICS-LAB/DNABERT_2}. The complete dataset composition, including sequence numbers, class distributions and sequence length statistics, is detailed in \cref{tab:GUE dataset}

\begin{table}[htbp]
\centering
\caption{The Composition of GUE Datasets}
\vskip 0.15in
\label{tab:GUE dataset}
\begin{tabular}{llccc}
\toprule
\textbf{Species} & \textbf{Task} & \textbf{Num. Datasets} & \textbf{Num. Classes} & \textbf{Sequence Length} \\ \midrule
\multirow{4}{*}{\textbf{Human}} 
 & Core Promoter Detection & 3 & 2 & 70 \\ 
 & Transcription Factor Prediction & 5 & 2 & 100 \\ 
 & Promoter Detection & 3 & 2 & 300 \\ 
 & Splice Site Detection & 1 & 3 & 400 \\ \midrule
\textbf{Mouse} & Transcription Factor Prediction & 5 & 2 & 100 \\ \midrule
\textbf{Yeast} & Epigenetic Marks Prediction & 10 & 2 & 500 \\ \midrule
\textbf{Virus} & Covid Variant Classification & 1 & 9 & 1000 \\ \bottomrule
\end{tabular}
\vskip -0.1in
\end{table}

\subsection{Implementation}

Building upon DNABERT2's downstream task hyperparameter framework, we systematically evaluated learning rates from {$5 \times 10^{-6}$, $5 \times 10^{-5}$, $6 \times 10^{-5}$, $7 \times 10^{-5}$, $8 \times 10^{-5}$, $3 \times 10^{-4}$} while maintaining a consistent batch size of 32 across all tasks. Task-specific learning rates were empirically determined through validation set performance. The optimization process employed the AdamW algorithm \citep{adamw} with 10,000 training steps, while retaining default parameter configurations from the HuggingFace Transformer Trainer implementation \citep{wolf-etal-2020-transformers}.

\begin{table}[H]
\caption{The results on the GUE datasets}
\vskip 0.15in
\begin{center}
\label{tab:GUE}
\resizebox{\textwidth}{!}{ % 缩放表格至页面宽度
\begin{tabular}{l*{6}{S[table-format=2.2]}}
\toprule

% 第一部分标题：Epigenetic Marks Prediction
% \multicolumn{7}{c}{\textbf{Epigenetic Marks Prediction}} \\
\cmidrule(lr){1-7}
\multirow{2}{*}{Model} & 
\multicolumn{6}{c}{\textbf{Epigenetic Marks Prediction}} \\
\cmidrule(lr){2-7}
& {H3} & {H3K14ac} & {H3K36me3} & {H3K4me1} & {H3K4me2} & {H3K4me3} \\
\cmidrule(lr){1-7}

% 第一部分数据
DNABERT (3-mer)        & 74.15 & 42.07 & 48.49  & 42.95  & 31.34  & 28.92  \\
DNABERT (4-mer)        & 73.03 & 41.88 & 48.03  & 41.06  & 30.66  & 25.31  \\
DNABERT (5-mer)        & 73.40 & 40.68 & 48.29  & 40.65  & 30.67  & 27.10  \\
DNABERT (6-mer)        & 73.10 & 40.06 & 47.25  & 41.44  & 32.27  & 27.81  \\
NT-500M-human          & 69.67 & 33.55 & 44.14  & 37.15  & 30.87  & 24.06  \\
NT-500M-1000g          & 72.52 & 39.37 & 45.58  & 40.45  & 31.05  & 26.16  \\
NT-2500M-1000g         & 74.61 & 44.08 & 50.86  & 43.10  & 30.28  & 30.87  \\
NT-2500M-multi         & 78.77 & \underline{56.20} & \textbf{61.99} & \textbf{55.30} & 36.49  & 40.34  \\
DNABERT-2              & 78.27 & 52.57 & 56.88  & 50.52  & 31.13  & 36.27  \\
DNABERT-2~$\blacksquare$ & \textbf{80.17} & \textbf{57.42} & \underline{61.90} & \underline{53.00} & \underline{39.89} & \underline{41.20} \\
\cmidrule(lr){1-7}
Enformer               & 70.65 & 37.87 & 42.41  & 34.00  & 29.65  & 22.19  \\
SPACE                   & \underline{79.53} & 54.12 & 54.82  & 50.92  & \textbf{43.80} & \textbf{49.47} \\
\cmidrule(lr){1-7}

\bottomrule
\end{tabular}}
\end{center}
\vskip -0.1in
\end{table}

\subsection{Results}

The results on the GUE datasets are presented in \cref{tab:GUE} and \cref{tab:GUE2}. In accordance with the implementation protocol of DNABERT2~\citep{DNABert2}, all benchmark tasks utilized the Matthews Correlation Coefficient (MCC) for performance evaluation, with the singular exception of viral sequence analysis where F1-score metrics were employed. The notation DNABERT2~$\blacksquare$ specifically denotes the model variant that underwent additional masked language modeling (MLM) pre-training on the training sets of the Genomic Understanding and Evaluation (GUE) benchmark, as detailed in the DNABERT2 methodology.

\begin{table}[H]
\raggedright 
\caption{The results on the GUE datasets.}
\vskip 0.15in
\ContinuedFloat % 续表标识
\label{tab:GUE2}
\begin{tabular}{l*{21}{S[table-format=2.2]}}  
\toprule

% 第二部分标题：Promoter Detection
% \multicolumn{7}{c}{\textbf{Promoter Detection}} \\
\cmidrule(lr){1-8}
\multirow{2}{*}{Model} & 
\multicolumn{4}{c}{\textbf{Epigenetic Marks Prediction}} &
\multicolumn{3}{c}{\textbf{Promoter Detection}}\\
\cmidrule(lr){2-8}
& {H3K79me3} & {H3K9ac} & {H4} & {H4ac} & {all} & {notata} & {tata} \\
\cmidrule(lr){1-8}

% 第二部分数据
DNABERT (3-mer)        & 60.12 & 50.48 & 78.27  & 38.60  & 90.44  & 93.61  & 69.83  \\
DNABERT (4-mer)        & 59.77 & 51.44 & 78.28  & 36.40  & 89.54  & 92.65  & 66.78  \\
DNABERT (5-mer)        & 59.61 & 51.11 & 77.27  & 37.48  & 90.16  & 92.45  & 69.51  \\
DNABERT (6-mer)        & 61.17 & 51.22 & 79.26  & 37.43  & 90.48  & 93.05  & 61.56  \\
NT-500M-human          & 58.35 & 45.81 & 76.17  & 33.74  & 87.71  & 90.75  & 78.07  \\
NT-500M-1000g          & 59.33 & 49.29 & 76.29  & 36.79  & 89.76  & 91.75  & 78.23  \\
NT-2500M-1000g         & 61.20 & 52.36 & 79.76  & 41.46  & 90.95  & 93.07  & 75.80  \\
NT-2500M-multi         & 64.70 & 56.01 & \underline{81.67}  & 49.13  & \underline{91.01} & 94.00  & \textbf{79.43} \\
DNABERT-2              & \textbf{67.39} & 55.63 & 80.71  & \underline{50.43} & 86.77  & \underline{94.27}  & 71.59  \\
DNABERT-2~$\blacksquare$ & 65.46 & \underline{57.07} & \textbf{81.86} & 50.35  & 88.31  & \textbf{94.34} & 68.79  \\

grover                 &    &   &   &   &86.42&92.3& 59.77 \\ 
\cmidrule(lr){1-8}
Enformer               & 55.69 & 49.35 & 76.32  & 32.90  & 85.68  & 92.92  & 69.63  \\
SPACE                   & \underline{66.93} & \textbf{59.29} & 81.25 & \textbf{53.09} & \textbf{91.90} & 94.23 & \underline{79.13} \\
\cmidrule(lr){1-8}

% 第三部分
\cmidrule(lr){1-9}
\multirow{2}{*}{Model} &
\multicolumn{5}{c}{\textbf{Transcription Factor Prediction (Human)}} &
\multicolumn{3}{c}{\textbf{Core Promoter Detection}} \\
\cmidrule(lr){2-9}
& \textbf{0} & \textbf{1} & \textbf{2} & \textbf{3} & \textbf{4} & \textbf{all} & \textbf{notata} & \textbf{tata} \\
\cmidrule(lr){1-9}

DNABERT(3-mer)        & 67.95 & 70.90 & 60.51 & 53.03 & 69.76 & \textbf{70.92} & 69.82 & \underline{78.15} \\
DNABERT(4-mer)        & 67.90 & 73.05 & 59.52 & 50.37 & 71.23 & 69.00 & 70.04 & 74.25 \\
DNABERT(5-mer)        & 66.97 & 69.98 & 59.03 & 52.95 & 69.26 & 69.48 & 69.81 & 76.79 \\
DNABERT(6-mer)        & 66.84 & 70.14 & 61.03 & 51.89 & 70.97 & 68.90 & \underline{70.47} & 76.06 \\
NT-500M-human         & 61.59 & 66.75 & 53.58 & 42.95 & 60.81 & 63.45 & 64.82 & 71.34 \\
NT-500M-1000g         & 63.64 & 70.17 & 52.73 & 45.24 & 62.82 & 66.70 & 67.17 & 73.52 \\
NT-2500M-1000g        & 66.31 & 68.30 & 58.70 & 49.08 & 67.59 & 67.39 & 67.46 & 69.66 \\
NT-2500M-multi        & 66.64 & 70.28 & 58.72 & 51.65 & 69.34 & \underline{70.33} & \textbf{71.58} & 72.97 \\
DNABERT-2             & \textbf{71.99} & \underline{76.06} & 66.52 & 58.54 & 77.43 & 69.37 & 68.04 & 74.17 \\
DNABERT-2~$\blacksquare$ & 69.12 & 71.87 & 62.96 & 55.35 & 74.94 & 67.50 & 69.53 & 76.18 \\

grover &65.76 & 67.9 &61.62 &48.26&74.68 &63.58&66.75&60.57\\

\cmidrule(lr){1-9}
Enformer              & \underline{69.42} & 72.76 & \textbf{77.88} & \textbf{66.41} & \underline{81.89} & 60.94 & 66.46 & 46.21 \\
SPACE                  & 69.02 & \textbf{76.49} & \underline{76.45} & \underline{66.08} & \textbf{82.91} & 68.18 & 68.04 & \textbf{79.23} \\
\cmidrule(lr){1-9}

% 第四部分
\cmidrule(lr){1-8}
\multirow{2}{*}{Model} &
\multicolumn{5}{c}{\textbf{Transcription Factor Prediction (Mouse)}} &
\multicolumn{1}{c}{\textbf{Virus}} &
\multicolumn{1}{c}{\textbf{Splice}} \\
\cmidrule(lr){2-8}
& \textbf{0} & \textbf{1} & \textbf{2} & \textbf{3} & \textbf{4} & \textbf{Covid} & \textbf{Splice} \\
\cmidrule(lr){1-8}

DNABERT(3-mer)        & 42.31 & 79.10 & 69.90 & 55.40 & 41.97 & 62.23 & 84.14 \\
DNABERT(4-mer)        & 49.42 & 79.95 & 72.62 & 51.79 & 44.13 & 59.87 & 84.05 \\
DNABERT(5-mer)        & 42.45 & 79.32 & 62.22 & 49.92 & 40.34 & 50.46 & 84.02 \\
DNABERT(6-mer)        & 44.42 & 78.94 & 71.44 & 44.89 & 42.48 & 55.50 & 84.07 \\
NT-500M-human         & 31.04 & 75.04 & 61.67 & 29.17 & 29.27 & 50.82 & 79.71 \\
NT-500M-1000g         & 39.26 & 75.49 & 64.70 & 33.07 & 34.01 & 52.06 & 80.97 \\
NT-2500M-1000g        & 48.31 & 80.02 & 70.14 & 42.25 & 43.40 & 66.73 & 85.78 \\
NT-2500M-multi        & 63.31 & 83.76 & 71.52 & 69.44 & 47.07 & \textbf{73.04} & \textbf{89.35} \\
DNABERT-2             & 56.76 & 84.77 & 79.32 & 66.47 & \textbf{52.66} & \underline{71.02} & 84.99 \\
DNABERT-2~$\blacksquare$ & 64.23 & \textbf{86.28} & 81.28 & \underline{73.49} & 50.80 & 68.49 & 85.93 \\

grover &&&&&&&84.35\\
\cmidrule(lr){1-8}
Enformer              & \textbf{67.15} & 81.56 & \underline{85.99} & 67.88 & 44.03 & 61.33 & 81.55 \\
SPACE                  & \underline{65.94} & \underline{84.91} & \textbf{90.30} & \textbf{86.72} & \underline{50.66} & 70.26 & \underline{87.48} \\
\cmidrule(lr){1-8}
\bottomrule
\end{tabular}
\vskip -0.1in
\end{table}

\section{Results on BEND Benchmark}
\begin{table}[H]
\centering
\caption{Results on all tasks of BEND}
\label{tab:bend}
\resizebox{\textwidth}{!}{ % 缩放表格至页面宽度
\begin{tabular}{lccccc}
\toprule

\multirow{2}{*}{\textbf{Method}} & \multicolumn{5}{c}{\textbf{Genomic Tasks}} \\
\cmidrule(lr){2-6}
& \textbf{Chromatin} & \textbf{Histone} & \textbf{CpG} & \textbf{Variant effects} & \textbf{Variant effects} \\
 & \textbf{accessibility} & \textbf{modification} & \textbf{Methylation} & \textbf{(expression)} & \textbf{(disease)} \\
\cmidrule(lr){1-6}
\textbf{Expert method} & 0.85 & 0.74 & 0.93 & 0.70 & 0.56 \\
&BASSET&BASSET&BASSET&DEEPSEA&DEEPSEA\\
\cmidrule(lr){1-6}
\textbf{Fully supervised} \\
\quad ResNet  & -- & -- & -- & -- & -- \\
\quad CNN  & 0.75 & 0.76 & 0.84 & -- & -- \\
\cmidrule(lr){1-6}
\textbf{Pre-trained} \\
\quad ResNet-LM  & 0.82 & 0.77 & 0.87 & 0.55 & 0.55 \\
\quad AWD-LSTM  & 0.69 & 0.74 & 0.81 & 0.53 & 0.45 \\
\quad NT-H  & 0.74 & 0.76 & 0.88 & 0.55 & 0.48 \\
\quad NT-MS  & 0.79 & 0.78 & \textbf{0.92} & 0.54 & 0.77 \\
\quad NT-1000G  & 0.77 & 0.77 & 0.89 & 0.45 & 0.49 \\
\quad NT-V2  & 0.80 & 0.76 & 0.91 & 0.48 & 0.48 \\
\quad DNABERT & 0.85 & 0.79 & 0.91 & \textbf{0.60} & 0.56 \\
\quad DNABERT-2 & 0.81 & 0.78 & 0.90 & 0.49 & 0.51 \\
\quad GENA-LM BERT & 0.76 & 0.78 & 0.91 & 0.49 & 0.55 \\
\quad GENA-LM BigBird & 0.82 & 0.78 & 0.91 & 0.49 & 0.52 \\
\quad HyenaDNA large & 0.84 & 0.76 & 0.91 & 0.51 & 0.45 \\
\quad HyenaDNA tiny& 0.78 & 0.76 & 0.86 & 0.47 & 0.44 \\
\quad GROVER & 0.82 & 0.77 & 0.89 & 0.56 & 0.51 \\
\quad GPN-MSA & -- & -- & -- & -- & \textbf{0.97} \\
\cmidrule(lr){1-6}
\quad SPACE &\textbf{0.89} &\textbf{0.81} &\textbf{0.92} &0.51 &0.49\\

\bottomrule
\end{tabular}
}
\end{table}
Experiments in the main paper involve complete fine-tuning of our Encoder parameters. In this section, we validate the effectiveness of SPACE's embeddings on the Bend Benchmark~\cite{marinbend} (i.e., directly using frozen SPACE embeddings for downstream tasks). All our experimental settings strictly follow the official configurations of Bend~\cite{marinbend}.
The results are shown in Figure~\ref{tab:bend}. We observe that SPACE achieves SOTA performance on chromatin accessibility, histone modification, and CpG methylation tasks, with chromatin accessibility surpassing the second-best method by 0.04. It is worth noting that although our supervised pre-training tasks also include chromatin accessibility and histone modification representations, which may provide potential advantages, their data processing approaches are not entirely identical. However, on the two variant effects tasks, SPACE shows limited effectiveness, similar to most DFMs. We hypothesize that masked language modeling tasks may be necessary to achieve good performance on variant effect prediction~\cite{benegas2025dna}.

\section{Genomic Benchmarks}
\subsection{Dataset}
Genomic Benchmarks currently comprises nine datasets focusing on regulatory elements (promoters, enhancers, and open chromatin regions) from three model organisms: Homo sapiens (human), Mus musculus (mouse), and Caenorhabditis elegans (nematode). All data were downloaded from \url{https://github.com/ML-Bioinfo-CEITEC/genomic_benchmarks}. The detailed composition of these datasets is presented in Table~\ref{genomic}.

\begin{table}[htbp]
\centering
\caption{Composition of Genomic Benchmarks.}
\vskip 0.15in
\label{genomic}
\begin{tabular}{lrrr}
\toprule
\textbf{Name} & \textbf{sequences} & \textbf{classes} & \textbf{Class ratio} \\
\midrule
dummy\_mouse\_enhancers\_ensembl & 1210 & 2 & 1.0 \\
demo\_coding\_vs\_intergenomic\_seqs & 100000 & 2 & 1.0 \\
demo\_human\_or\_worm & 100000 & 2 & 1.0 \\
drosophila\_enhancers\_stark & 6914 & 2 & 1.0 \\
human\_enhancers\_cohn & 27791 & 2 & 1.0 \\
human\_enhancers\_ensembl & 154842 & 2 & 1.0 \\
human\_ensembl\_regulatory & 289061 & 3 & 1.2 \\
human\_nontata\_promoters & 36131 & 2 & 1.2 \\
human\_ocr\_ensembl & 174756 & 2 & 1.0 \\
\bottomrule
\end{tabular}
\end{table}
\subsection{Implementation}
We systematically evaluated learning rates {$5 \times 10^{-6}$, $5 \times 10^{-5}$, $6 \times 10^{-5}$, $7 \times 10^{-5}$, $8 \times 10^{-5}$, $3 \times 10^{-4}$} and batch sizes {8, 16, 32, 64}. The optimal learning rate and batch size for each task were determined through validation set performance experiments. The optimization process employed the AdamW algorithm \citep{adamw} with 3 training epochs, while maintaining the default parameter configuration from the HuggingFace Transformer Trainer implementation \citep{wolf-etal-2020-transformers}.

\section{Ablation Study}
\label{appendix:ablation}

SPACE demonstrates comparable or superior performance to the decoder-removed variant in 14/18 tasks, with 11/18 tasks still outperforming even when replaced by a parameter-matched MLP. Notably, for regulatory element classification tasks, SPACE achieves better results in 4/5 datasets, with the only exception being the TATA box dataset—which primarily examines sequence motifs of TATA boxes and does not require complex regulatory mechanism understanding. This suggests that while our decoder does not explicitly improve direct chromatin profile prediction accuracy, the MoE architecture implicitly captures cross-profile regulatory interactions by modeling their dependencies. This capability provides critical advantages for tasks requiring integrated understanding of multiple profiles, such as regulatory element prediction.

\begin{table}[htbp]
\caption{Ablation study on NT downstream tasks.}
\vskip 0.15in
\centering
\label{tab:ablation_NT}
\resizebox{\textwidth}{!}{
\begin{tabular}{l*{6}{S[table-format=1.3]}}
\toprule

\cmidrule(lr){1-7}
\multirow{2}{*}{Model} & 
\multicolumn{6}{c}{\textbf{Chromatin profiles}} \\
\cmidrule(lr){2-7}
& {H2AFZ} & {H3K27ac} & {H3K27me3} & {H3K36me3} & {H3K4me1} & {H3K4me2} \\
\cmidrule(lr){1-7}
% \textbf{Model with hidden dimensions halved}\\
\quad SPACE w/o decoder                       & 0.535 & 0.514 & 0.567 & 0.593 & 0.520 & 0.604 \\
\quad SPACE w/o decoder w/ MLP                 & 0.551 & 0.528 & 0.577 & 0.580 & 0.534 & 0.637 \\           
\quad SPACE w/o encoder                       & 0.540 & 0.524 & 0.569 & 0.579 & 0.506 & 0.625 \\
\quad SPACE w/o encoder and species emb         & 0.551 & 0.518 & 0.566 & 0.585 & 0.519 & 0.622 \\
\quad SPACE                                 & 0.556 & 0.529 & 0.579 & 0.593 & 0.516 & 0.612 \\
\cmidrule(lr){1-7}
% \textbf{Model with full parameters}\\
% \quad Enformer                              & 0.522 & 0.520 & 0.552 & 0.567 & 0.504 & 0.626 \\
% \quad SPACE w/o species embedding           & 0.544 & 0.545 & 0.586 & 0.608 & 0.544 & 0.639 \\
% \quad SPACE random emb and gate             & 0.549 & 0.539 & 0.585 & 0.601 & 0.545 & 0.634 \\ 
% \quad SPACE                                 & 0.548 & 0.547 & 0.586 & 0.602 & 0.543 & 0.640 \\
% \cmidrule(lr){1-7}

\cmidrule(lr){1-7}
\multirow{2}{*}{Model} & 
\multicolumn{4}{c}{\textbf{Chromatin profiles}}&
\multicolumn{2}{c}{\textbf{Regulatory elements}}\\
\cmidrule(lr){2-7}
& {H3K4me3} & {H3K9ac} & {H3K9me3} & {H4K20me1} & {Enhancers} & {Enhancers(types)} \\
\cmidrule(lr){1-7}
% \textbf{Model with hidden dimensions halved}\\
\quad SPACE w/o decoder                       & 0.661 & 0.601 & 0.452 & 0.627 & 0.598 & 0.563 \\
\quad SPACE w/o decoder w/ MLP                 & 0.668 & 0.589 & 0.451 & 0.636 & 0.601 & 0.558 \\           
\quad SPACE w/o encoder                       & 0.627 & 0.585 & 0.461 & 0.637 & 0.612 & 0.564 \\
\quad SPACE w/o encoder and species emb         & 0.654 & 0.588 & 0.454 & 0.635 & 0.596 & 0.563 \\
\quad SPACE                                 & 0.637 & 0.582 & 0.457 & 0.644 & 0.607 & 0.564 \\
\cmidrule(lr){1-7}

\cmidrule(lr){1-7}
\multirow{2}{*}{Model} & 
\multicolumn{3}{c}{\textbf{Regulatory elements}} &
\multicolumn{3}{c}{\textbf{Splicing}}\\
\cmidrule(lr){2-7}
& {All} & {NoTATA} & {TATA} & {Acceptors} & {All} & {Donors} \\
\cmidrule(lr){1-7}
% \textbf{Model with hidden dimensions halved}\\
\quad SPACE w/o decoder                       & 0.752 & 0.773 & 0.841 & 0.873 & 0.884 & 0.936 \\
\quad SPACE w/o decoder w/ MLP                 & 0.743 & 0.750 & 0.808 & 0.883 & 0.886 & 0.937\\           
\quad SPACE w/o encoder                       & 0.738 & 0.769 & 0.828 & 0.864 & 0.869 & 0.933 \\
\quad SPACE w/o encoder and species emb         & 0.739 & 0.767 & 0.828 & 0.869 & 0.876 & 0.942 \\
\quad SPACE                                 & 0.763 & 0.776 & 0.802 & 0.898 & 0.884 & 0.941 \\
\cmidrule(lr){1-7}

\bottomrule
\end{tabular}}
\vskip -0.1in
\end{table}

\begin{table}[htbp]
\caption{Ablation study on GUE benchmarks.}
\vskip 0.15in
\begin{center}
\label{tab:ablation_GUE}
% 第一个表格用更小的字体并居中
\resizebox{\textwidth}{!}{
\begin{tabular}{l*{5}{S[table-format=2.2]}}
\toprule

\multirow{2}{*}{\small Model} & 
\multicolumn{5}{c}{\small\textbf{Epigenetic Marks Prediction}}\\
\cmidrule(lr){2-6}
& {\small H3} & {\small H3K14ac} & {\small H3K36me3} & {\small H3K4me1} & {\small H3K4me2} \\ 
\midrule
{\small SPACE w/o dec}      &{\small 76.76} &{\small 46.75} &{\small 50.09} &{\small 39.56} &{\small 34.80} \\
{\small SPACE w/o dec w/ MLP}  & {\small 75.59} & {\small 45.17} & {\small 48.21}  & {\small 39.70}  & {\small 34.81}  \\
{\small SPACE w/o enc}      & {\small 76.16} & {\small 48.78} & {\small 49.14}  & {\small 37.57}  & {\small 34.08}  \\
{\small SPACE w/o enc and species emb} &{\small 76.94} &{\small 48.77} &{\small 42.46} &{\small 43.01} &{\small 34.33}\\
{\small SPACE}          & {\small 76.40} & {\small 50.76} & {\small 49.18}  & {\small 41.30}  & {\small 32.83}  \\
\bottomrule
\end{tabular}}

\vspace{0.1in}

\resizebox{\textwidth}{!}{
\begin{tabular}{l*{6}{S[table-format=2.2]}}
\toprule
\multirow{2}{*}{Model} & 
\multicolumn{5}{c}{\textbf{Epigenetic Marks Prediction}}& 
\multicolumn{1}{c}{\textbf{Virus}}\\
\cmidrule(lr){2-6} \cmidrule(lr){7-7}
& {H3K4me3} & {H3K79me3} & {H3K9ac} & {H4} & {H4ac} &{Covid} \\ 
\midrule
SPACE w/o dec           &{34.85} &{57.85} &{55.38} &{79.78} &{49.05} &{68.66} \\
SPACE w/o dec w/ MLP   & {34.26}  & {58.94} & {56.36} & {78.81} & {43.49} & {67.83} \\
SPACE w/o enc       & {36.84}  & {63.44} & {56.63} & {77.17} & {50.78} & {68.46} \\
SPACE w/o enc and species emb &{37.13} &{63.84}  &{56.27} &{78.29} &{51.14} &{68.56}\\
SPACE           & {37.74}  & {61.10} & {57.06} & {79.33} & {51.05} & {68.89} \\

\bottomrule
\end{tabular}}

\end{center}
\vskip -0.1in
\end{table}

\section{Model Parameter Counts}
\label{appendix:parameter}

We present the parameter counts of SPACE and its ablation variants in Table~\ref{tab:parameter}. The SPACE (large) configuration represents our primary model with complete architectural components for comparative analysis, while the other variants correspond to reduced-scale models specifically designed for ablation studies. These smaller models employ 131 KB input sequences with a compressed hidden dimension of 768 and operate under a batch size of 32.

\begin{table}[H]
\centering
\caption{Model Parameter Counts of SPACE and its ablation variants}
\vskip 0.15in
\label{tab:parameter}
\begin{tabular}{lcccc}
\toprule
 & SPACE (large) & SPACE w/o enhancement & SPACE w/o species MoE & SPACE (small) \\
\midrule
param counts &588.75M &150.96M & 105.19M &183.19M\\
hidden dim  & 1536 &768 &768 &768 \\
\bottomrule
\end{tabular}
\vskip -0.1in
\end{table}

It should be particularly noted that, based on the sparse architecture design of the MoE, our model activates only a partial subset of parameters during a single forward computation. This selective parameter activation mechanism makes the number of effective parameters actually involved in the computation significantly lower than the total number of parameters in the model, thus significantly reducing the computational resource consumption while maintaining the model capacity.

\end{document}